\pdfoutput=1
\PassOptionsToPackage{table,dvipsnames}{xcolor}%
\documentclass[10pt,twocolumn,letterpaper]{article}

\usepackage[pagenumbers]{cvpr} %

\usepackage{amssymb}
\usepackage{multirow}
\usepackage{amsmath}
\usepackage{pifont}
\usepackage[dvipsnames]{xcolor}
\newcommand{\cmark}{\ding{51}}%
\newcommand{\xmark}{\ding{55}}%
\usepackage{tikz}

\usepackage[accsupp]{axessibility}  %

\newcommand{\third}{\cellcolor{ForestGreen!20}}
\newcommand{\scnd}{\cellcolor{ForestGreen!40}}
\newcommand{\frst}{\cellcolor{ForestGreen!75}}

\newcommand{\NH}{\includegraphics[height=0.8cm]{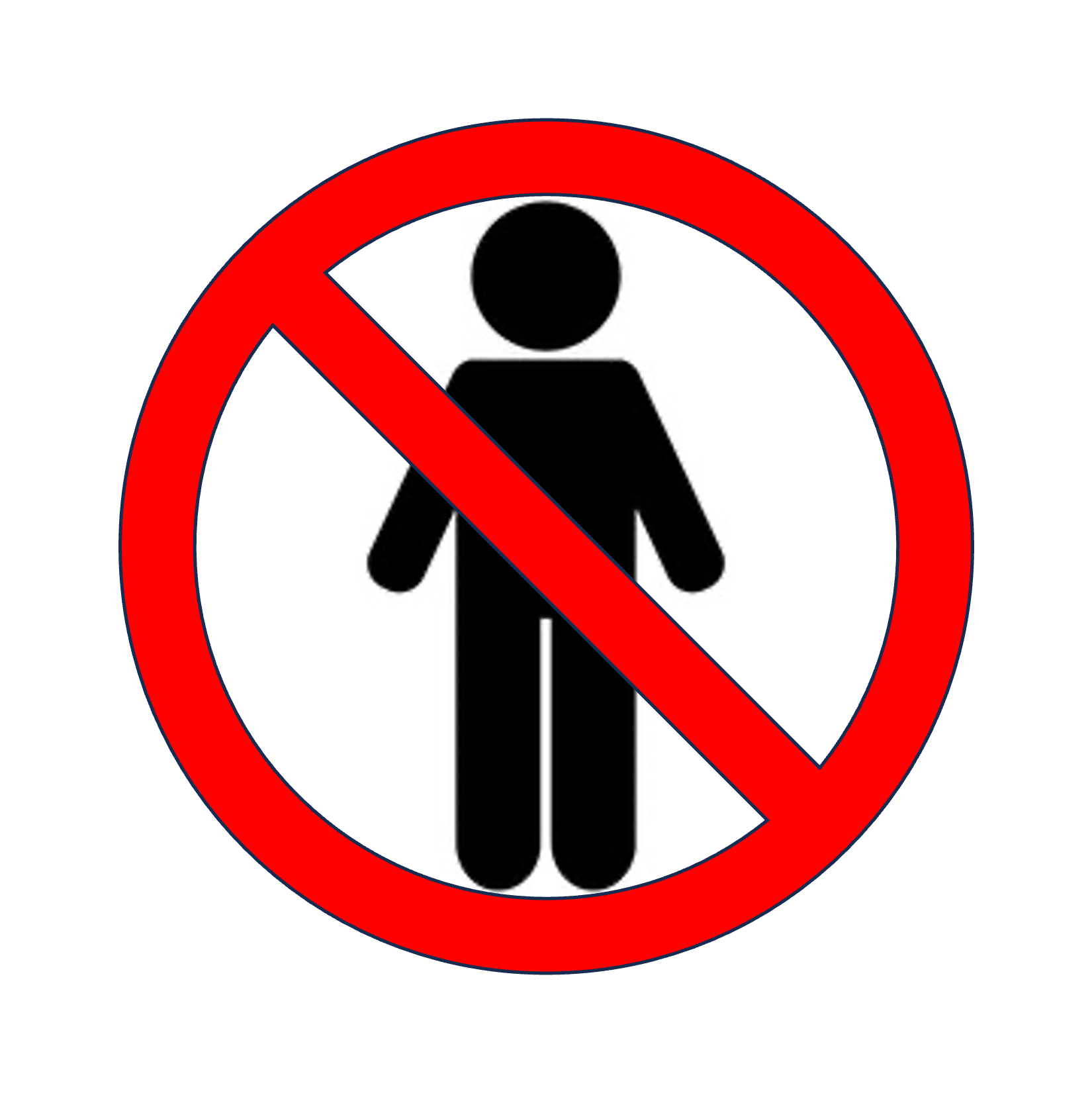}}

\newcommand{\greencheck}{{\color{Green}\cmark}}
\newcommand{\redcross}{{\color{red}\xmark}}

\newcommand\blfootnote[1]{%
  \begingroup
  \renewcommand\thefootnote{}\footnote{#1}%
  \addtocounter{footnote}{-1}%
  \endgroup
}

\definecolor{cvprblue}{rgb}{0.21,0.49,0.74}
\usepackage[pagebackref,breaklinks,colorlinks,allcolors=cvprblue]{hyperref}

\def\paperID{16729} %
\def\confName{CVPR}
\def\confYear{2025}

\title{Augmenting Perceptual Super-Resolution via Image Quality Predictors}

\author{
Fengjia Zhang*
\hspace{9mm}
Samrudhdhi B. Rangrej*
\hspace{9mm}
Tristan Aumentado-Armstrong*
\\
Afsaneh Fazly 
\hspace{9mm}
Alex Levinshtein
\\
{ AI Center -- Toronto, Samsung Electronics}\\
{\tt\small \{f.zhang2, s.rangrej, tristan.a, a.fazly, alex.lev\}@samsung.com}
}

\begin{document}

\twocolumn[{
\maketitle
\begin{center}
    \captionsetup{type=figure}
    \includegraphics[width=\textwidth]{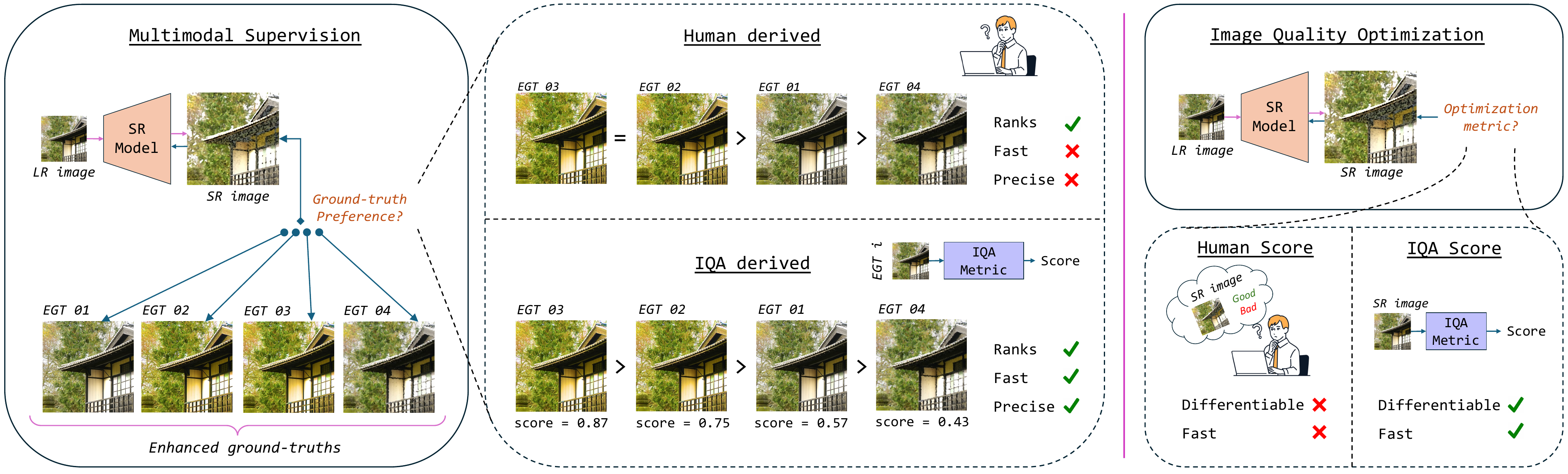}
    \captionof{figure}{\textbf{Schematics for improving perceptual super-resolution (SR).} Perceptual quality of SR can be improved in two ways: \textbf{(Left)} providing supervision through multiple enhanced ground-truths (EGT) or \textbf{(Right)} direct optimization for the quality of the super-resolved image. In both cases, human-in-the-loop can greatly improve performance. However, manual annotation is tedious, imprecise, and non-differentiable. An IQA metric can replace a human in rating the enhanced ground-truths or can directly act as a differentiable optimization objective. In this paper, we specifically assess whether more practical no-reference (NR) IQA metrics can replace human raters for SR. We find that combining NR-IQA-based sampling and regularized optimization is sufficient to attain state-of-the-art perceptual image quality, \textit{without} requiring human ratings.
    }
    \label{fig:teaser}
\end{center}
}]

\begin{abstract}
Super-resolution (SR), a classical inverse problem in computer vision, is inherently ill-posed, inducing a distribution of plausible solutions for every input. However, the desired result is not simply the expectation of this distribution, which is the blurry image obtained by minimizing pixelwise error, but rather the sample with the highest image quality. A variety of techniques, from perceptual metrics to adversarial losses, are employed to this end. In this work, we explore an alternative: utilizing powerful non-reference image quality assessment (NR-IQA) models in the SR context. We begin with a comprehensive analysis of NR-IQA metrics on human-derived SR data, identifying both the accuracy (human alignment) and complementarity of different metrics. Then, we explore two methods of applying NR-IQA models to SR learning: (i) altering data sampling, by building on an existing multi-ground-truth SR framework, and (ii) directly optimizing a differentiable quality score. Our results demonstrate a more human-centric perception-distortion tradeoff, focusing less on non-perceptual pixelwise distortion, instead improving the balance between perceptual fidelity and human-tuned NR-IQA measures.
\end{abstract}
    
\blfootnote{*Equal Contribution.}
\vspace{-0.5cm}\section{Introduction}
\label{sec:intro}

Many tasks in human and computer vision are naturally formulated as ill-posed inverse problems \cite{pizlo2001perception}.
Single-image super-resolution (SISR),
which has many practical applications to digital photographic zoom, 
is a well-studied example of this 
(e.g., \cite{baker2002limits,freeman2002example}).
In SISR, a given low-resolution (LR) image has an associated distribution of high-resolution (HR) ``real'' images that could have given rise to it. 
The fundamental challenge of SR is therefore not just to find \textit{any} sample from that distribution, but instead to find perceptually plausible one(s).
Early learning-based models, trained with pixel-wise losses (e.g., \cite{dong2014learning,dong2016accelerating}), effectively ``average'' over possible solutions in pixel-space, resulting in blurry output images with a high peak signal-to-noise ratio (PSNR).
However, human preferences indicate that a solution with high image quality is better than this averaged one.
Hence, numerous techniques have been devised to emphasize perceptual fidelity, 
	such as perceptual metrics \cite{johnson2016perceptual,ding2021comparison} and adversarial losses (e.g., \cite{saxena2021generative}), greatly improving image quality.
In other words, pixelwise fidelity is a poor measure of perceptual quality.
In fact, under some conditions, they are directly oppositional, 
	forming a ``perception--distortion tradeoff'' \cite{blau2018perception,blau2019rethinking}.
In theory, the only pixel-space constraint is given by the LR image; 
	besides this, the optimal SR result (in terms of human preference) may have very high pixelwise distortion (w.r.t. the ``real'' ground-truth generating image), as long as it has high plausibility with respect to the LR input and high image quality.

Instead of optimizing pixel-space distortions, 
	we focus on improving perceptual image quality. This is commonly done using a combination of perceptual losses and GANs \cite{ledig2017photo, wang2018esrgan, pan2021exploiting, umer2020deep, ma2020structure}, enabling the SR model to target a multi-modal distribution rather than specific ground-truth targets. The challenge of such methods, however, is to produce perceptually plausible outputs without introducing high-frequency artifacts \cite{liang2022details}. Many full-reference (FR) \cite{kettunen2019lpips,zhang2018unreasonable,ghazanfari2023r,ding2020image}  and  non-reference image quality assessment (NR-IQA) \cite{saad2010dct,tang2011learning,talebi2018nima,ke2021musiq} metrics were developed to align with human preferences for identifying perceptually plausible images. While some approaches started to replace perceptual losses with FR metrics like LPIPS \cite{kettunen2019lpips} and DISTS \cite{ding2020image}, for the task of image restoration, NR-IQA metrics are still used purely for evaluation purposes. Motivated by the adoption of human feedback guidance in text-to-image generative models (e.g., \cite{xu2024imagereward,fan2023dpok,clark2023directly,prabhudesai2023aligning}), we aim to use NR-IQA metrics to improve SISR.

 Recently, Chen et al. \cite{chen2023human} used human feedback to improve SISR. They do so by generating multiple enhanced versions of GT (Fig.~\ref{fig:teaser} left), manually rating these different versions using multiple human evaluators, and fine-tuning the model on the positively ranked GTs. While this results in significant perceptual quality improvements without introducing unwanted artifacts, manual human ranking is very coarse and cumbersome. We instead use an automatic NR-IQA measure that is well-correlated with human scores, yielding a more fine-grained ranking, and bypassing the requirement for having human feedback (Fig.~\ref{fig:teaser} centre). Additionally, since the measure is fully differentiable, it can be used for direct optimization (e.g., replacing or complementing GANs), unlike human scores that cannot be used in this fashion (Fig.~\ref{fig:teaser} right). Our contributions are as follows:

\begin{itemize}
\item 
We present a detailed analysis of NR-IQA on two human-derived SR datasets, thus identifying metrics that are generally useful for improving SR image quality. 
\item 
We explore the application of NR-IQA to SISR, via two approaches: sampling across multiple GTs weighted by NR-IQA, and direct optimization of NR-IQA. 
\item 
We achieve SISR results that are perceptually on par or better than SISR finetuned with human feedback, but using an automatic NR-IQA measure instead. 
\end{itemize}

\begin{table*}[t]
    \centering
    \resizebox{\linewidth}{!}{%
    \begin{tabular}{c | c c c c c c c}
    \toprule
        Method & PaQ-2-PiQ \cite{ying2020patches} & NIMA$^\dagger$ \cite{talebi2018nima} & MUSIQ$^\triangleleft$ \cite{ke2021musiq} & LIQE$^\heartsuit$ \cite{zhang2023blind} & ARNIQA-TID$^\star$ \cite{agnolucci2024arniqa} & Q-Align$^\diamond$ \cite{wu2024qalign} & TOPIQ-NR \cite{chen2024topiq} \\
    \midrule
        Accuracy (\%) & 76.41 & 74.91 & 74.47 & 74.03 & 74.03 & 73.77 & 73.06 \\
    \bottomrule
    \end{tabular}
    }
    \caption{\textbf{Phase I analysis on SBS180K.} Accuracy of top 7 NR-IQA metrics on a subset (1212 image pairs) from train portion. We use the default configuration of the metrics, provided by the \texttt{IQA-PyTorch} toolbox, unless stated otherwise. $^\dagger$We use NIMA with Inception V2 as base model. $^\triangleleft$We use the default MUSIQ trained on KonIQ~\cite{hosu2020koniq}. $^\heartsuit$We use LIQE pretrained on KonIQ~\cite{hosu2020koniq}. $^\star$We use ARNIQA metric trained on TID2013~\cite{ponomarenko2013color}. $^\diamond$We use Q-Align metric specialized in image quality assessment.}
    \label{tab:analysis_phaseI}
\end{table*}

\begin{table*}[t]
    \centering
    \resizebox{\linewidth}{!}{%
    \begin{tabular}{c | c c c c c c c}
    \toprule
        Method & PaQ-2-PiQ \cite{ying2020patches} & NIMA$^\dagger$ \cite{talebi2018nima} & MUSIQ$^\triangleleft$ \cite{ke2021musiq} & LIQE$^\heartsuit$ \cite{zhang2023blind} & ARNIQA-TID$^\star$ \cite{agnolucci2024arniqa} & Q-Align$^\diamond$ \cite{wu2024qalign} & TOPIQ-NR \cite{chen2024topiq} \\
    \midrule
        Train Acc. (\%) & 80.41 & 79.32 & 79.96 & 77.70 & 77.74 & 80.00 & 78.30 \\
        Test Acc. (\%) & 80.57 & 81.37 & 82.73 & 77.45 & 77.07 & 80.68 & 81.28 \\
    \bottomrule
    \end{tabular}
    }
    \caption{\textbf{Phase II analysis on SBS180K dataset.} Accuracy of top 7 (according to Phase I) NR-IQA metrics on the entire train and test sets of SBS180K. We exclude image pairs where humans prefer both images equally. $^{\dagger\triangleleft\heartsuit\star\diamond}$ See Table~\ref{tab:analysis_phaseI}.}
    \label{tab:analysis_phaseII}
\end{table*}

\section{Related Work}
\label{sec:relwork}

{\bf Deep learning-based SISR:} 
Deep learning has given a significant boost to SISR performance, taking the SOTA mantle from dictionary-based methods \cite{timofte2013anchored, timofte2015a+, yang2012coupled} to CNN-based approaches like SRNet and its successors \cite{dong2014learning, dong2016accelerating, kim2016deeply, ahn2018fast}. Since then, there came many architectural improvements: deeper architectures, like RCAN \cite{zhang2018image}, hierarchical processing \cite{lai2017deep}, advanced building blocks like in NAFNet \cite{chen2022simple} and RRDB from ESRGAN \cite{wang2018esrgan}, and better upsampling like PixelUnshuffle \cite{shi2016real}, to name a few. Due to long range dependencies in SISR, transformer-based methods \cite{liang2021swinir, chen2023activating}, and auto-regressive models based on Mamba \cite{liu2024vmamba}, have recently achieved SOTA performance \cite{chen2024ntire2024challengeimage,guo2024mambair}. 

\noindent
{\bf Perceptual quality-oriented SISR:} 
Early SISR approaches optimized a simple pixel-wise reconstruction loss, such as L2 or L1, between model output and ground-truth \cite{dong2014learning,dong2016accelerating}. However, due to the ill-posedness of SISR, this yields poor perceptual quality. Blau and Michaeli \cite{blau2019rethinking} have shown that there exists a tradeoff between good perceptual quality and accurate reconstruction (or fidelity). To improve perceptual quality, perceptual losses were proposed, such as SSIM \cite{wang2004image}, which measures patch similarity rather than per-pixel similarity, and later others \cite{johnson2016perceptual} that measure the similarity between deep VGG features rather than pixel intensities. Combined with perceptual losses, GAN-based training \cite{goodfellow2014generative} was used to improve perceptual quality in SRGAN \cite{ledig2017photo} and many follow-up approaches \cite{wang2018esrgan, pan2021exploiting, umer2020deep, ma2020structure, bahat2020explorable}. One challenge of such approaches is to improve perceptual quality without introducing unwanted hallucinations. Proposed solutions include more specialized discriminators \cite{park2023content} and better balancing between various loss terms \cite{liang2022details, park2023perception}. Rather than changing the training loss function, perceptual quality can be improved by explicitly generating multiple training targets \cite{jo2021tackling, chen2023human}, or encouraged using specific architectural designs. Normalizing flows \cite{lugmayr2020srflow,yao2023local} were used to directly output a distribution over plausible solutions rather than a single SR image. More recently, diffusion-based approaches \cite{saharia2022image, kawar2022denoising, xia2023diffir, wang2024exploiting,wu2024seesr,wu2024one,yang2024pixel} and alike \cite{delbracio2023inversion} have been shown to achieve a better perceptual quality than GAN-based methods.

\noindent
{\bf Human guided perceptual quality assessment:}
One may use image quality assessment (IQA) metrics to evaluate the 
aesthetic quality of super-resolved images. They can be full-reference (FR) or no-reference (NR) metrics. In real-world applications where the true HR image is unavailable, NR-IQA metrics are more useful.
Early opinion-unaware NR-IQA metrics used hand-crafted features to assess how closely the statistics of the output images match with natural scene statistics, e.g., BRISQUE~\cite{mittal2012no} and NIQE~\cite{zhang2015feature}. Others developed metrics that align with human preferences \cite{saad2010dct,tang2011learning}. %
Recently, many developed deep-learning-based opinion-unaware approaches such as FID~\cite{heusel2017gans}, and opinion-aware approaches such as NIMA~\cite{talebi2018nima} and MUSIQ~\cite{ke2021musiq}. 
While the above metrics are general purpose, metrics such as NRQM~\cite{ma2017learning} and NeuralSBS~\cite{khrulkov2021neural} are specifically designed to align with human preferences for super-resolution.
Recently, human preferences are being widely incorporated in improving generative models, especially for text-to-image generation~\cite{xu2024imagereward, liang2024rich, clark2023directly}. Yet, incorporating human guidance in improving SR models has not received much attention. Ding \etal~\cite{ding2021comparison} attempt to use various \textit{full-reference} IQA metrics for SR model finetuning, and Chen \etal~\cite{chen2023human} propose to use \textit{human guidance in GT selection process}.
Unlike these works, we attempt to assess the ability of \textit{no-reference} IQA metrics to (a) \textit{automatically} rate and select optimal GT, eliminating the need for arduous manual annotation, and (b) act as a finetuning objective to improve aesthetic quality of super-resolved images.

\section{Analysis of NR-IQA metrics}
\label{sec:analysis}

We analyze the alignment between various NR-IQA metrics and human judgements for assessing the aesthetic quality of super-resolved images. We report our analysis on two publicly available datasets, SBS180K~\cite{khrulkov2021neural} and HGGT~\cite{chen2023human}.

\subsection{Analysis on the SBS180K dataset}
\label{analysis:sbs}

We analyze various NR-IQA metrics on SBS180K~\cite{khrulkov2021neural}, a large scale human preference dataset for super-resolved images, containing 167,019 train and 9,421 test image pairs. Each pair is annotated with a single score in the range [0,1], depicting the fraction of human annotators preferring the aesthetic quality of the second image over the first one. Each pair consists of two super-resolved versions of the same low-resolution image, with each version generated using a different SR model.
Due to the large size of SBS180K, we analyze the metrics in two phases. In phase I, we analyze 20 NR-IQA metrics and their variants (total 42 metrics) on a small subset of the train set. In phase II, we analyze the top 7 NR-IQA metrics from phase I on the complete train and test sets. We use \texttt{IQA-PyTorch}\cite{pyiqa}, an open-source toolbox for image quality assessment.

\noindent \textbf{Phase I.} There are 404 unique pairs of compared SR models in the train set. We randomly select 3 image pairs per model comparison, yielding a subset of 1212 images. We evaluate the following NR-IQA metrics on this subset: Q-Align \cite{wu2024qalign}, LIQE \cite{zhang2023blind}, ARNIQA \cite{agnolucci2024arniqa}, TOPIQ \cite{chen2024topiq}, TReS \cite{golestaneh2022no}, CLIPIQA(+) \cite{wang2023exploring}, MANIQA \cite{yang2022maniqa}, MUSIQ \cite{ke2021musiq}, DBCNN \cite{zhang2020blind}, PaQ-2-PiQ \cite{ying2020patches}, HyperIQA \cite{su2020blindly}, NIMA \cite{talebi2018nima}, WaDIQaM \cite{bosse2017deep}, CNNIQA \cite{zheng2021learning}, NRQM \cite{ma2017learning}, PI (Perceptual Index) \cite{blau20182018}, BRISQUE \cite{mittal2012no}, ILNIQE and NIQE \cite{zhang2015feature}, and PIQE \cite{venkatanath2015blind}. When available, 
we also consider multiple variants of these metrics offered by \texttt{IQA-PyTorch} (e.g., metrics trained on different IQA datasets).
We assess the accuracy of each metric in terms of whether, given an image-pair, the metric prefers the same image as the humans prefer collectively or not.
Table \ref{tab:analysis_phaseI} shows results of only the top 7 metrics: PaQ-2-PiQ, NIMA, MUSIQ, LIQE, ARNIQA, Q-Align, TOPIQ-NR. Complete results are given in Supp.\ Table~\ref{tab:Phase I_supp}.

\noindent \textbf{Phase II.} We evaluate the top 7 metrics from Phase I on the complete train and test sets, excluding pairs with no consensus among human annotators (score of 0.5). Results are given in Table~\ref{tab:analysis_phaseII}, suggesting that MUSIQ has relatively higher accuracy on both train and test sets compared to other metrics. While PaQ-2-PiQ and Q-Align have slightly higher accuracy (0.45\% and 0.04\%, respectively) on the train set, MUSIQ outperforms them on the test set by a large margin (2.16\% and 2.05\%, respectively).
We further analyze performance of the remaining six metrics on the samples where MUSIQ fails (excluding pairs with a score of 0.5). %
Results are shown in Table~\ref{tab:analysis_phaseII_musiq_failure_case}. We find that NIMA and Q-Align achieve higher accuracy on these samples compared to other four metrics. Since they complement MUSIQ, in \S\ref{sec:results}, we report performance on MUSIQ, NIMA and Q-Align.
\begin{table*}[t]
    \centering
    \begin{tabular}{c | c c c c c c}
    \toprule
        Method & PaQ-2-PiQ \cite{ying2020patches} & NIMA$^\dagger$ \cite{talebi2018nima} & LIQE$^\heartsuit$ \cite{zhang2023blind} & ARNIQA-TID$^\star$ \cite{agnolucci2024arniqa} & Q-Align$^\diamond$ \cite{wu2024qalign} & TOPIQ-NR \cite{chen2024topiq} \\
    \midrule
        Train Acc. (\%) & 37.58 & 44.21 & 31.50 & 42.05 & 42.86 & 25.71 \\
        Test Acc. (\%) & 33.10 & 39.70 & 30.70 & 38.60 & 41.06 & 30.89 \\
    \bottomrule
    \end{tabular}
    \caption{\textbf{Phase II analysis on SBS180K (continued).} Accuracy of top 6 NR-IQA metrics on consensus samples (from train plus test) where MUSIQ fails. %
    $^{\dagger\heartsuit\star\diamond}$ See Table~\ref{tab:analysis_phaseI}.}
    \label{tab:analysis_phaseII_musiq_failure_case}
\end{table*}
\begin{table*}[t]
    \centering
    \resizebox{\linewidth}{!}{%
    \begin{tabular}{c | c c c c c c c}
    \toprule
        Method & PaQ-2-PiQ \cite{ying2020patches} & NIMA$^\dagger$ \cite{talebi2018nima} & MUSIQ$^\triangleleft$ \cite{ke2021musiq} & LIQE$^\heartsuit$ \cite{zhang2023blind} & ARNIQA-TID$^\star$ \cite{agnolucci2024arniqa} & Q-Align$^\diamond$ \cite{wu2024qalign} & TOPIQ-NR \cite{chen2024topiq} \\
    \midrule
        SRC $\uparrow$ & 0.10 & 0.17 & 0.17 & 0.03 & 0.28 & 0.20 & 0.09 \\
        PM $\downarrow$ & 0.26 & 0.28 & 0.16 & 0.58 & 0.51 & 0.39 & 0.21 \\
        NM $\downarrow$ & 0.85 & 0.79 & 0.95 & 0.64 & 0.45 & 0.63 & 0.97 \\
    \bottomrule
    \end{tabular}
    }
    \caption{\textbf{Analysis on HGGT subset.}  We evaluate Spearman's rank correlation coefficient (SRC), positive misalignment (PM) rate, and negative misalignment (NM) rate. $^{\dagger\triangleleft\heartsuit\star\diamond}$. See Table~\ref{tab:analysis_phaseI}.
    }
    \label{tab:analysis_HGGT}
\end{table*}

\subsection{Analysis on HGGT dataset}
\label{analysis:hggt}

We analyze the seven selected metrics of Phase I above on the HGGT~\cite{chen2023human} dataset, containing 20,193 quintuplets of HR image patches. Each quintuplet contains an original HR ground-truth (GT) patch and four enhanced GTs. Each of the four enhanced GTs in each quintuplet is annotated by human annotators for being better than (`positive'), similar to (`similar'), or worse than (`negative') the original GT. While `positive' labels are abundant, `negative' labels are rare. Out of 20,193, only 1,270 quintuplets have at least one `negative'. We analyze the seven metrics on this subset.

We evaluate the NR-IQA metrics based on the average Spearman rank correlation coefficient, and positive and negative misalignment rates. Assuming higher rank is better, we define positive (negative) misalignment rate as the fraction of quintuplets where at least one positive (negative) GT is ranked lower (higher) than at least one similar GT. We show results in Table~\ref{tab:analysis_HGGT}. Note that all metrics have poor negative misalignment rate, leading to low Spearman correlations. We believe that NR-IQA metrics fail to recognize negative GTs, since they may not necessarily have low quality (recall that all are enhanced GTs), or may have artifacts that are unrecognizable without a reference image. Nonetheless, MUSIQ has the lowest positive misalignment rate. Hence, we use MUSIQ in \S\ref{sec:methods} for weighted sampling of the GTs and direct optimization (see also Supp.~\S\ref{supp:sec:nriqachoices}). Since TOPIQ has the 2nd lowest positive misalignment rate after MUSIQ, we include it as an evaluation metric in \S\ref{sec:results}.

\section{Methods}
\label{sec:methods}

We next explore how to improve existing SR methods with the results of our findings.
Since our interest is in \textit{perceptual} quality and its use in multimodal SR, we build upon recent work, Human Guided Ground-truth (HGGT) \cite{chen2023human}, which constructs a \textit{set} of ground-truth images per input, with varying quality, and uses human tests to rank their relative quality.
We begin by reviewing HGGT \cite{chen2023human} (\S\ref{sec:methods:bg}), and then discuss two methods of applying neural IQA models to augment it: 
altering the choice of ground-truth set based on an automated IQA weight (\S\ref{sec:methods:rs}) 
and 
directly optimizing the IQA model in a fine-tuning step (\S\ref{sec:methods:do}) .

\subsection{Background}
\label{sec:methods:bg}

As discussed in \S\ref{analysis:hggt}, the HGGT dataset includes
(i) a set of images (``originals''),
(ii) a set of four super-resolved versions of each original 
    (``enhanced GTs''), and
(iii) human annotations for each enhanced GT 
    (``positive'', ``similar'', or ``negative'' meaning better than, indistinguishable from, or worse than the original).
The set of positives provides \textit{multimodal} supervision,
    since each one is a disparate yet reasonable GT for learning.
The HGGT work \cite{chen2023human} then shows that utilizing these synthetic GTs is useful for SR training, exploring several neural architectures and degradation settings.
While HGGT explores several variants for utilizing their human labels, we focus on the simple but highly performing ``positives-only'' scenario, which performs equivalently or better than the variants utilizing negatives.
In this scenario, at each training iteration, every input image supervises the network with a GT chosen \textit{uniformly randomly} from the positives.
As is relatively standard in SR (e.g., \cite{wang2021real,ji2020real}), HGGT models are trained with a combined loss:
\begin{equation}
\label{eq:loss}
    \mathcal{L}(\theta|\widehat{I},I) = 
    \lambda_{\ell_1} || I - \widehat{I} ||_1 +
    \lambda_P d_\mathrm{P}(\widehat{I},I) +
    \lambda_A D(\widehat{I}),
\end{equation}
where $\widehat{I} = f_\theta(I_{\mathrm{LQ}})$ is the SR estimate of the low-resolution (or low-quality) input $I_{\mathrm{LQ}}$, via SR network $f_\theta$, 
$I \sim\mathcal{U}_{I_{\mathrm{LQ}}}[\{I_1,\ldots,I_n\}]$
is the randomly chosen GT (from the set of positives corresponding to image $I_{\mathrm{LQ}}$),
$d_\mathrm{P}$ is a perceptual loss,
and $D$ is an adversarial discriminator.

However, HGGT requires human labels, which are difficult to scale and often domain-dependent. 
In contrast, we explore the opportunities afforded by neural no-reference image quality assessment (NR-IQA) models, which not only eschew human labels, but also confer additional capabilities -- namely, the ability to provide more fine-grained non-uniform sampling weights (\S\ref{sec:methods:rs}) and to enable direct optimization via differentiability (\S\ref{sec:methods:do}).

\begin{figure}
    \centering
    \begin{minipage}[t]{0.32\linewidth}
    \includegraphics[width=\linewidth]{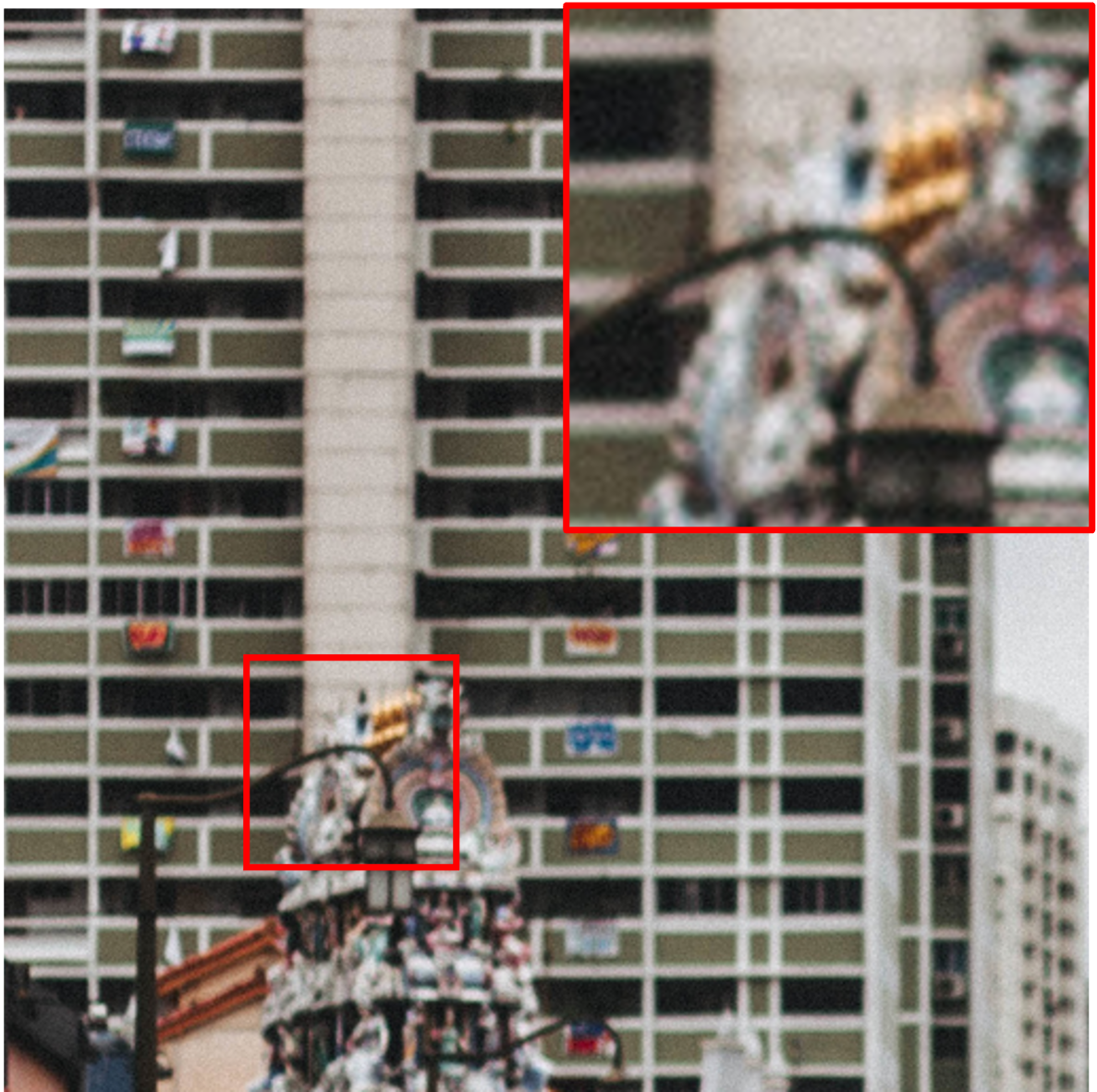}
    \centering{Original GT}
    \centering{MUSIQ = 31.76}
    \end{minipage}
    \hfill
    \begin{minipage}[t]{0.32\linewidth}
    \includegraphics[width=\linewidth]{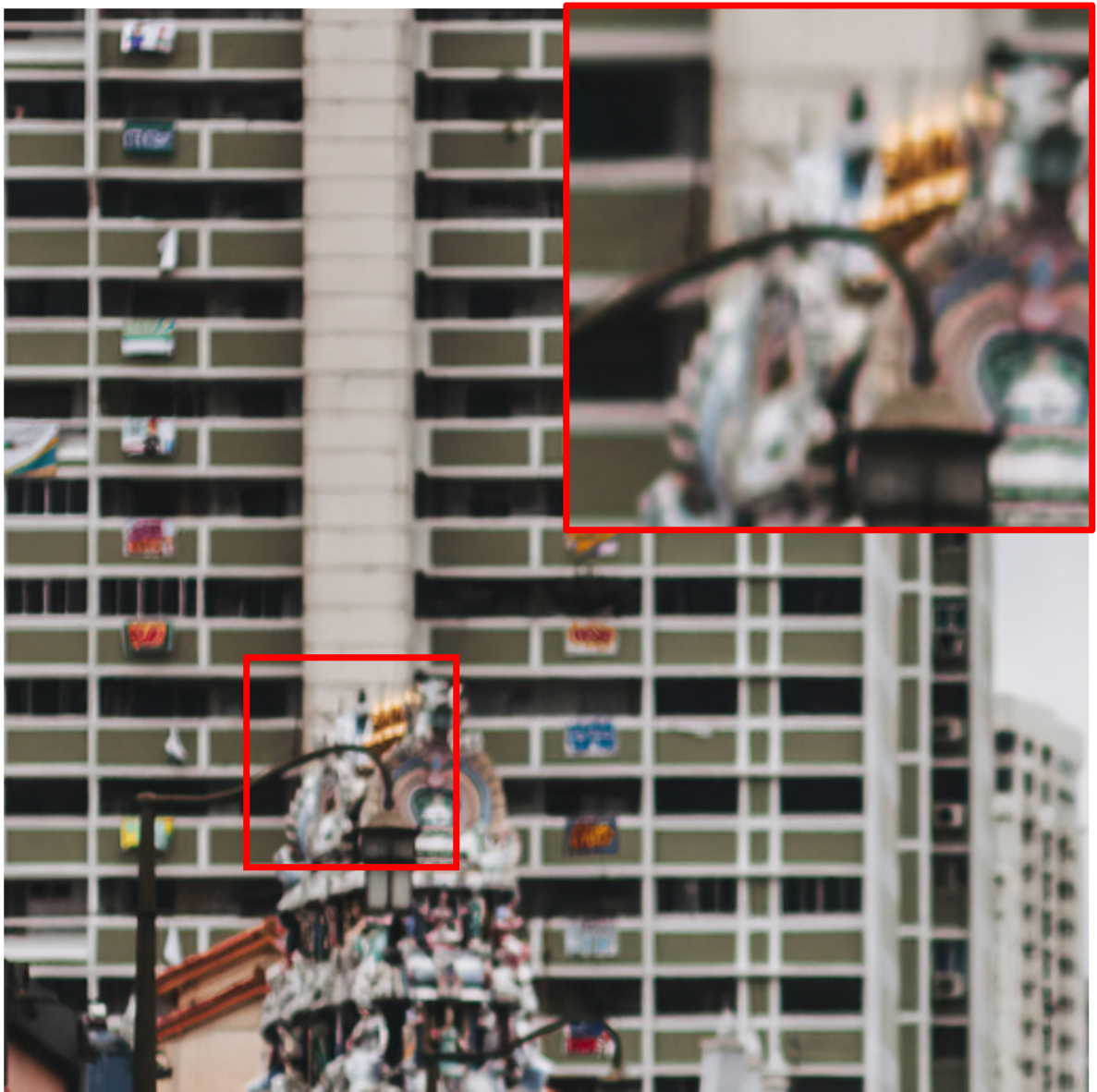}
    \centering{Positive GT 1}
    \centering{MUSIQ = 36.13}
    \end{minipage}
    \hfill
    \begin{minipage}[t]{0.32\linewidth}
    \includegraphics[width=\linewidth]{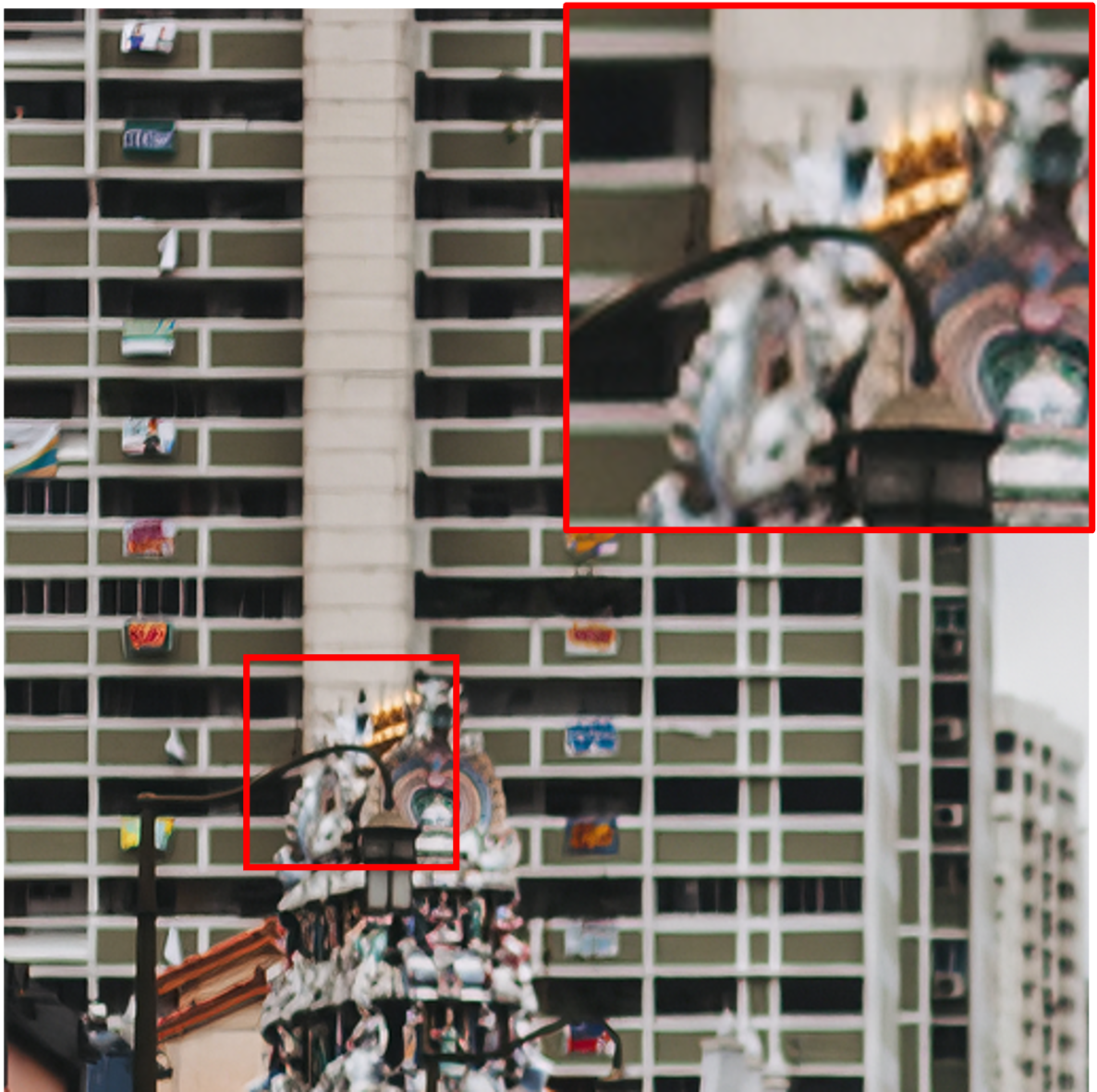}
    \centering{Positive GT 2}
    \centering{MUSIQ = 54.19}
    \end{minipage}
    \caption{
        \textbf{Fine-Grained Comparison via NR-IQA.}
        MUSIQ can differentiate the quality of two images, both marked as ``positive'' by human annotators.
        Higher MUSIQ indicates higher quality (zoom for details).
        Unlike HGGT models, which utilize a uniform distribution over positives, our approach enables differently weighting them
        (\S\ref{sec:methods:rs}).
    }
    \label{fig:musiq_for_positive}
\end{figure}

\subsection{Reweighted Sampling}
\label{sec:methods:rs}

We explore a few straightforward alternatives to uniform sampling of the positives, via an IQA model. In particular, consider the following simple formulation:
\begin{align}
    I &\sim \mathcal{P}[ S_I \mid \mathrm{SoftMax}_\tau(Q(S_I)) ], \\ \label{eq:gt_sample}
    Q(S_I) &= \{Q(I_1),\ldots,Q(I_n)\}, \\
    S_I &= \{I_1,\ldots,I_n\}\in \{ A_I, P_I \}
\end{align}
where $I$ is the sampled GT, 
$\tau > 0$ is the softmax temperature, 
$Q$ is the NR-IQA model (higher is better), 
$\mathcal{P}$ is a discrete distribution over elements of $S_I$ (weighted by $\mathrm{SoftMax}_\tau(Q(S_I))$), and
$S_I$ is the set of possible GTs (either choosing from \textit{all} candidates, enhanced and original, denoted $A_I$, or just \textit{positive} ones, $P_I$).
The HGGT algorithm simply uses $S_I = P_I$ and $\tau \rightarrow\infty$ (i.e., the uniform distribution); we explore different combinations, including $\tau\rightarrow 0$ (the $\arg\max$ choice).
We illustrate the utility of NR-IQA-based sampling (as opposed to uniform) in
Fig.~\ref{fig:musiq_for_positive}, displaying an example that humans rank equivalently as positive, yet is more precisely distinguished by the neural assessor.
We consider three NR-IQA-based sampling scenarios.

\noindent
\textbf{Softmax-All (SMA).}
Given the set of all GTs (i.e., $S_I = A_I$), 
we use an IQA-weighted distribution over GTs.
This setting uses no human data, and simply randomly chooses a GT at each iteration with a weight proportional to softmax-rescaled quality.
We set $\tau$ to ensure a distribution between uniform and Kronecker delta (i.e., argmax).

\noindent
\textbf{Softmax-Positives (SMP).}
This approach actually builds on the human data in HGGT, using the softmax-normalized IQA scores but only of the positives
(i.e., $S_I = P_I$).
This setting is the most similar to the HGGT positives-only (or uniform distribution on positives), just with non-uniform weights (based on $\tau$).
We expect this to outperform SMA sampling, as it has access to direct human preferences.

\noindent
\textbf{Argmax-online (AMO).}
The use of a neural IQA model confers an additional capability that human data lacks: we can dynamically determine sampling weights for new patches at training time. 
In previous scenarios, at training time, we first pick one GT out of the four (Eq. \ref{eq:gt_sample}), followed by random patch sampling from the selected GT.
Instead, in the Argmax-online (AMO) scenario, we first sample a random patch from each GT, followed by selecting the best patch.
To be specific, we sample one patch from the same random location from each GT, then run $Q$ on each patch, and choose the best one 
(i.e., the $\arg\max$ of $Q$ values, so
$\tau\rightarrow 0$).
We eschew human data; hence, $S_I = A_I$.
This enables a more fine-grained judgment (since quality is computed at the \textit{patch} level), whereas human annotations cannot necessarily be so easily extrapolated.

\subsection{Direct Optimization}
\label{sec:methods:do}

\begin{figure}
    \centering 
    \includegraphics[width=0.495\linewidth]{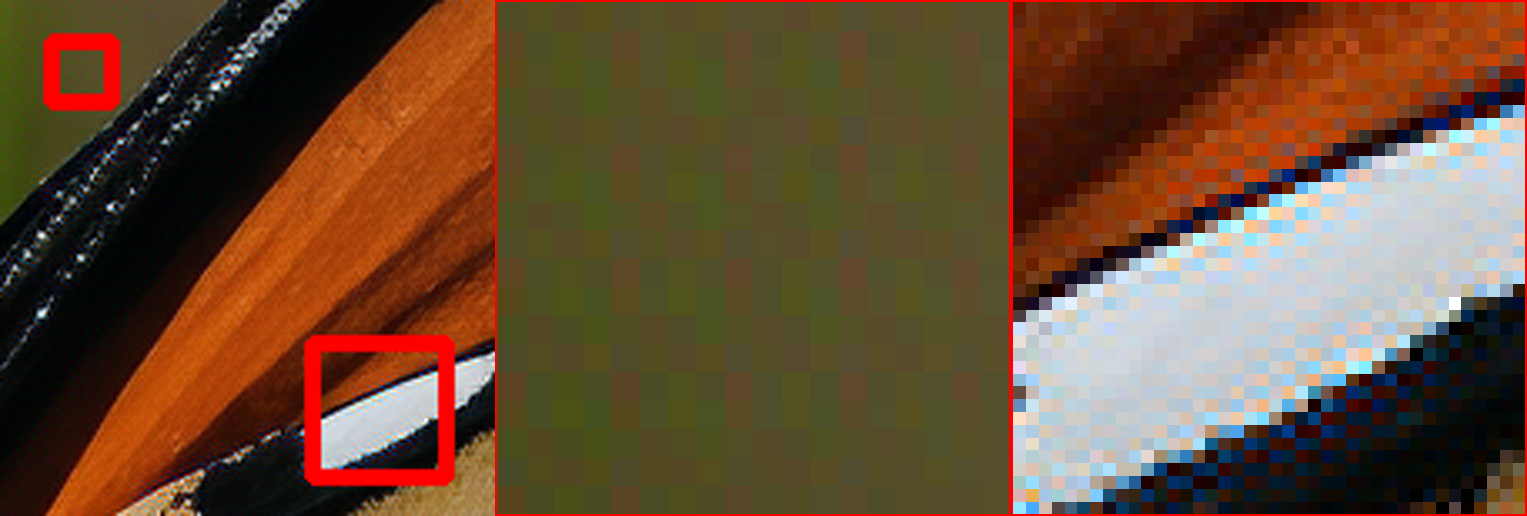}\includegraphics[width=0.495\linewidth]{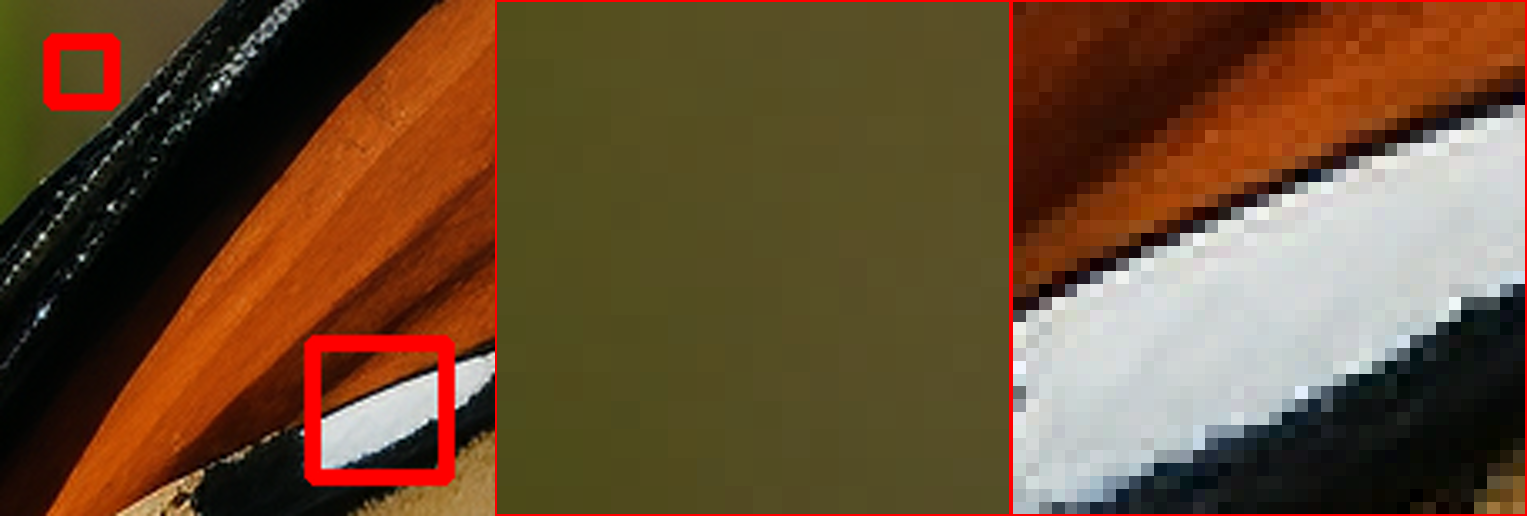}
    \caption{
        \textbf{Structured Optimization Noise.}
        Optimizing via an NR-IQA metric (MUSIQ \cite{ke2021musiq}) generates structured artifacts (left), similar to an adversarial attack, while utilizing LoRA removes this noise (right; see \S\ref{sec:methods:do} and Supp.~\S\ref{supp:sec:optart}). Zoom in for details.
    }
    \vspace{-0.25cm}\label{fig:advartifacts}
\end{figure}

Given a differentiable image quality estimator, $Q$, an obvious approach to improving our SR model is to simply include $Q$ in our objective function.
To some extent, this has already been explored for unconditional generative models (e.g., \cite{clark2023directly}).
However, when $Q$ is a neural network with many parameters, this is unlikely to succeed; in essence, gradient descent will act like an ``adversarial attack'' on $Q$ (e.g., \cite{goodfellow2014explaining}).
It is well-known that such ``attacks'' are often able to dramatically alter the output of the objective network (say, a classifier), 
while changing the optimized input in unintuitive or imperceptible ways
(e.g., \cite{kurakin2016adversarial,szegedy2014intriguing});
in SR, this could conceivably manifest as artifacts that fool $Q$ into providing a high score, since NR-IQA models are known to be susceptible to attacks (e.g., \cite{gushchin2024guardians,liu2024defense}).
In fact, without additional regularization, this is precisely what happens:
in Fig.~\ref{fig:advartifacts}, we display the artifacts that appear when an SR network is naively fine-tuned with $Q$ (see also Supp.~\S\ref{supp:sec:optart}).

Thus, inspired by prior work \cite{clark2023directly}, 
we utilize low-rank adaptation (LoRA) \cite{hu2021lora} to regularize the optimization.
Formally, we continue training as normal, but only on the LoRA weights, plus an additional NR-IQA loss term:
\begin{equation}
    \widetilde{\mathcal{L}}(\phi|\widehat{I},I) = 
    \mathcal{L}(\phi|\widehat{I},I)
    - \lambda_Q Q(\widehat{I}),
\end{equation}
where $\phi$ are the LoRA parameters,
$\widehat{I} = f_{\theta,\phi}(I_{\mathrm{LQ}})$,
$Q$ is an NR-IQA model (where higher is better), and $\mathcal{L}$ is defined in 
Eq.~\ref{eq:loss}.
Unless otherwise specified, we set $\lambda_A = 0$ when fine-tuning, since we are already including an image quality term (for which the critic normally acts; though see \S\ref{sec:ablations}).
See Supp.\ \S\ref{supp:methods} for additional details.

\section{Experiments}
\label{sec:results}

\begin{figure*}
    \centering
    \begin{minipage}[t]{0.19\linewidth}
        \includegraphics[width=\linewidth]{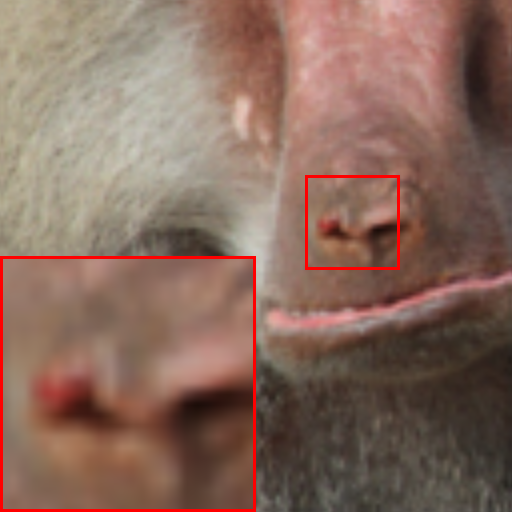}
        \vspace{-1.25cm}
        \begin{center}
        {\hspace{2.25cm}\colorbox{gray}{\color{white}24.42}}
        \end{center}
    \end{minipage}
    \begin{minipage}[t]{0.19\linewidth}
        \includegraphics[width=\linewidth]{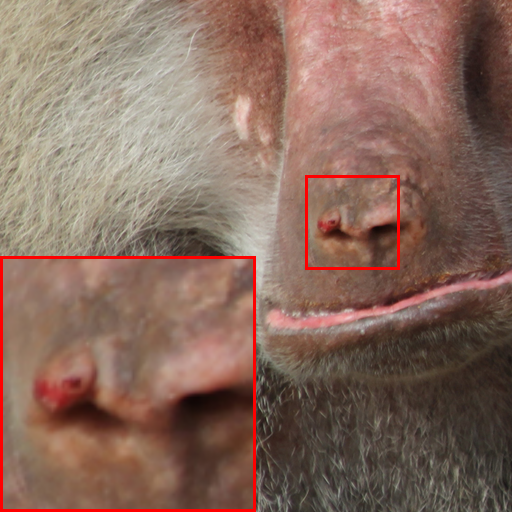}
        \vspace{-1.25cm}
        \begin{center}
        {\hspace{2.25cm}\colorbox{gray}{\color{white}56.62}}
        \end{center}
    \end{minipage}
    \begin{minipage}[t]{0.19\linewidth}
        \includegraphics[width=\linewidth]{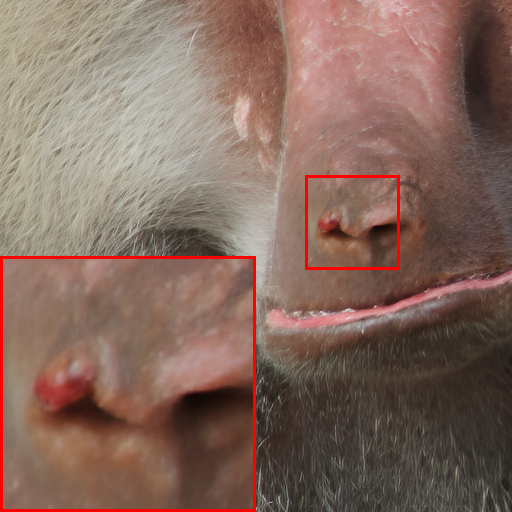}
        \vspace{-1.25cm}
        \begin{center}
        {\hspace{2.25cm}\colorbox{gray}{\color{white}60.86}}
        \end{center}
    \end{minipage}
    \begin{minipage}[t]{0.19\linewidth}
        \includegraphics[width=\linewidth]{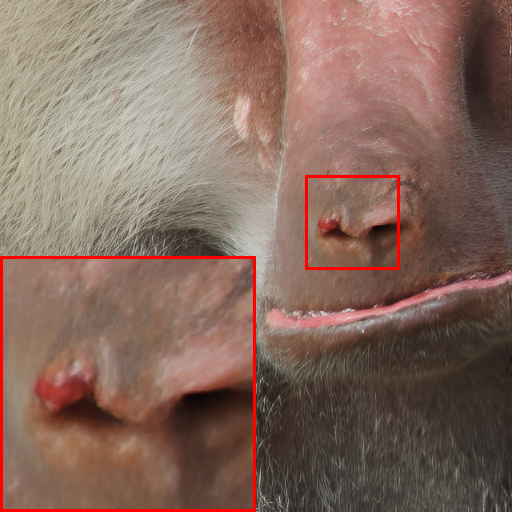}
        \vspace{-1.25cm}
        \begin{center}
        {\hspace{2.25cm}\colorbox{gray}{\color{white}65.14}}
        \end{center}
    \end{minipage}
    \begin{minipage}[t]{0.19\linewidth}
        \includegraphics[width=\linewidth]{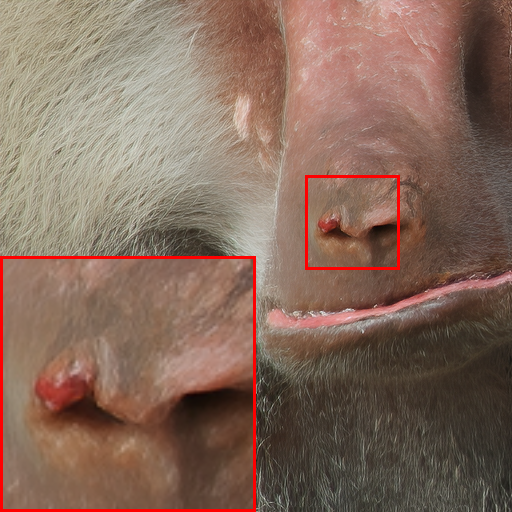}
        \vspace{-1.25cm}
        \begin{center}
        {\hspace{2.25cm}\colorbox{gray}{\color{white}73.12}}
        \end{center}
    \end{minipage}
    \vfill
    \begin{minipage}[t]{0.19\linewidth}
        \includegraphics[width=\linewidth]{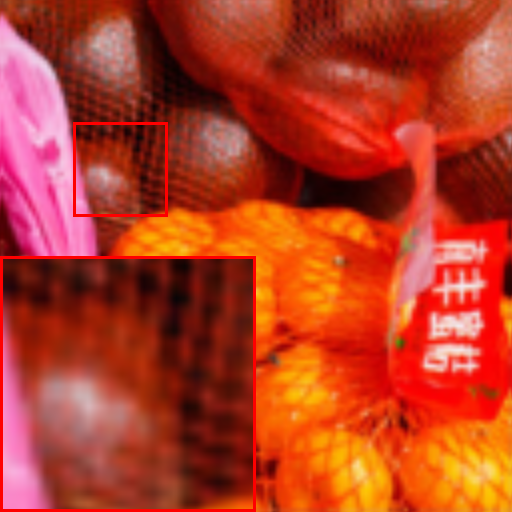}
        \vspace{-1.25cm}
        \begin{center}
        {\hspace{2.25cm}\colorbox{gray}{\color{white}47.59}}
        \end{center}
    \end{minipage}
    \begin{minipage}[t]{0.19\linewidth}
        \includegraphics[width=\linewidth]{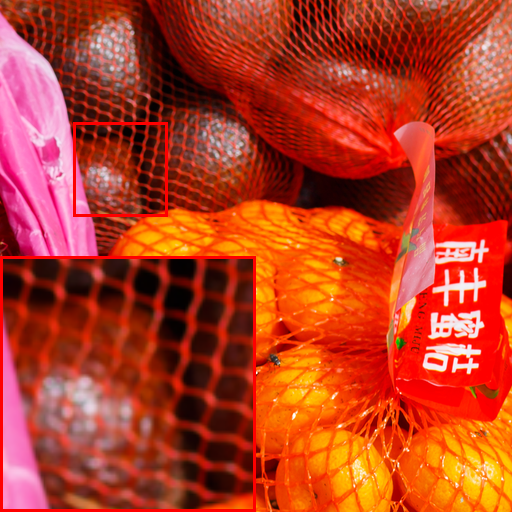}
        \vspace{-1.25cm}
        \begin{center}
        {\hspace{2.25cm}\colorbox{gray}{\color{white}71.13}}
        \end{center}
    \end{minipage}
    \begin{minipage}[t]{0.19\linewidth}
        \includegraphics[width=\linewidth]{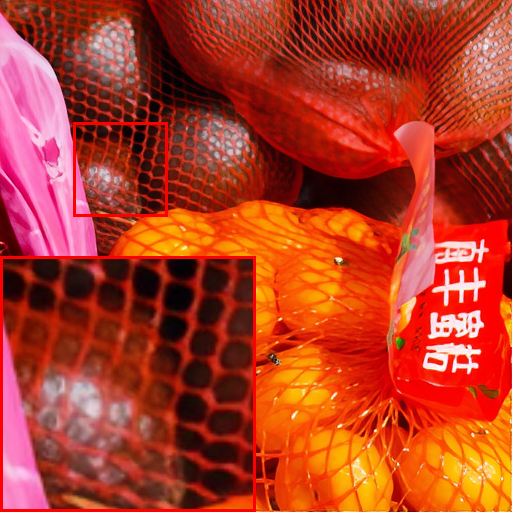}
        \vspace{-1.25cm}
        \begin{center}
        {\hspace{2.25cm}\colorbox{gray}{\color{white}74.13}}
        \end{center}
    \end{minipage}
    \begin{minipage}[t]{0.19\linewidth}
        \includegraphics[width=\linewidth]{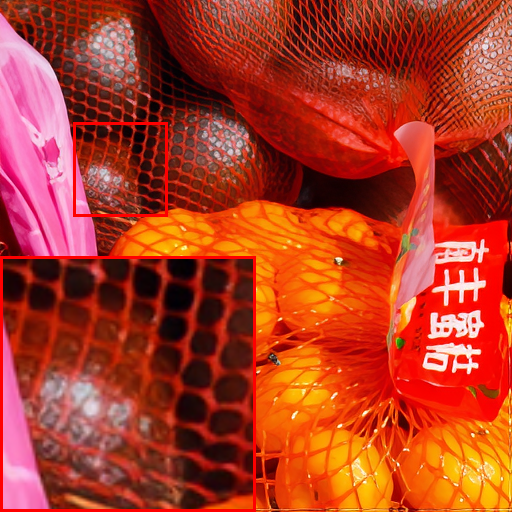}
        \vspace{-1.25cm}
        \begin{center}
        {\hspace{2.25cm}\colorbox{gray}{\color{white}75.78}}
        \end{center}
    \end{minipage}
    \begin{minipage}[t]{0.19\linewidth}
        \includegraphics[width=\linewidth]{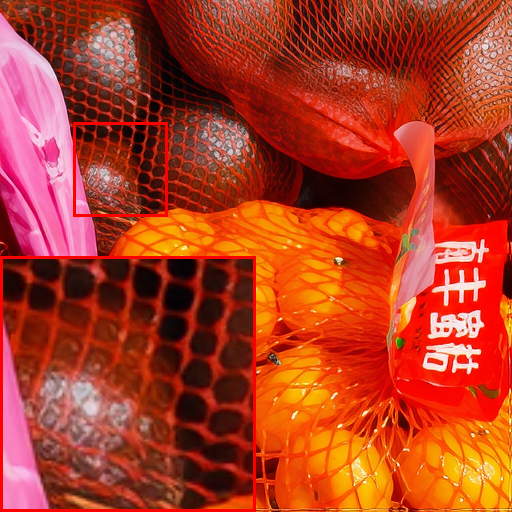}
        \vspace{-1.25cm}
        \begin{center}
        {\hspace{2.25cm}\colorbox{gray}{\color{white}77.31}}
        \end{center}
    \end{minipage}
    \vfill
    \begin{minipage}[t]{0.19\linewidth}
    \centering{LR}
    \end{minipage}%
    \begin{minipage}[t]{0.20\linewidth}
    \centering{Original GT}
    \end{minipage}%
    \begin{minipage}[t]{0.19\linewidth}
    \centering{SwinIR-UPos}
    \end{minipage}%
    \begin{minipage}[t]{0.20\linewidth}
    \centering{SwinIR-AMO}
    \end{minipage}%
    \begin{minipage}[t]{0.19\linewidth}
    \centering{SwinIR-AMO+FT}
    \end{minipage}
    \vfill
    \begin{minipage}[t]{0.19\linewidth}
        \includegraphics[width=\linewidth]{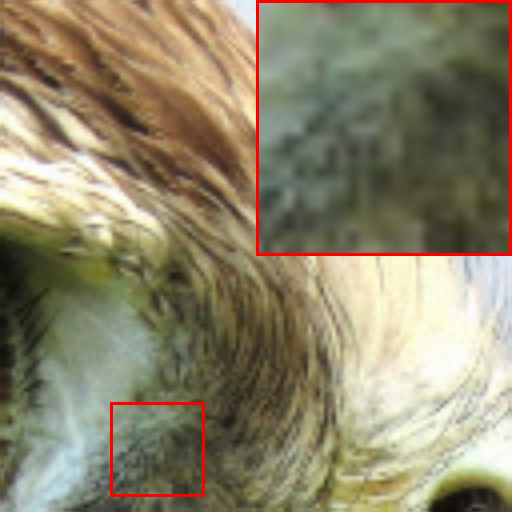}
        \vspace{-1.25cm}
        \begin{center}
            {\hspace{2.25cm}\colorbox{gray}{\color{white}41.46}}
        \end{center}
    \end{minipage}
    \begin{minipage}[t]{0.19\linewidth}
        \includegraphics[width=\linewidth]{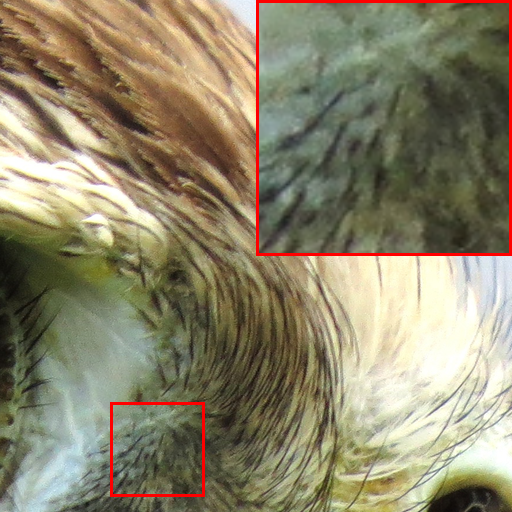}
        \vspace{-1.25cm}
        \begin{center}
        {\hspace{2.25cm}\colorbox{gray}{\color{white}52.51}}
        \end{center}
    \end{minipage}
    \begin{minipage}[t]{0.19\linewidth}
        \includegraphics[width=\linewidth]{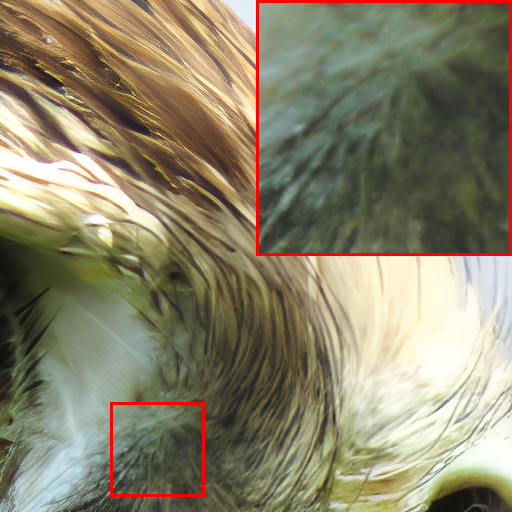}
        \vspace{-1.25cm}
        \begin{center}
        {\hspace{2.25cm}\colorbox{gray}{\color{white}54.92}}
        \end{center}
    \end{minipage}
    \begin{minipage}[t]{0.19\linewidth}
        \includegraphics[width=\linewidth]{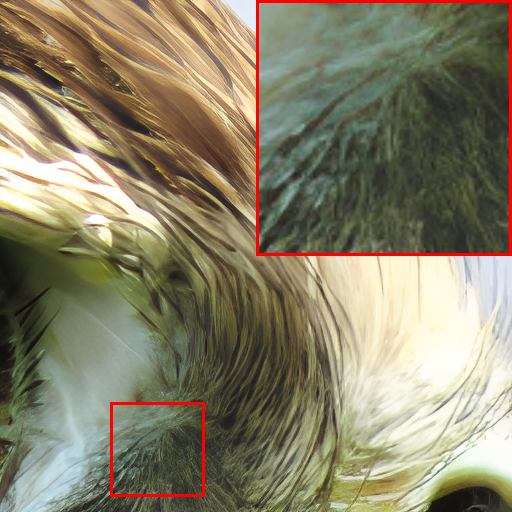}
        \vspace{-1.25cm}
        \begin{center}
        {\hspace{2.25cm}\colorbox{gray}{\color{white}66.12}}
        \end{center}
    \end{minipage}
    \begin{minipage}[t]{0.19\linewidth}
        \includegraphics[width=\linewidth]{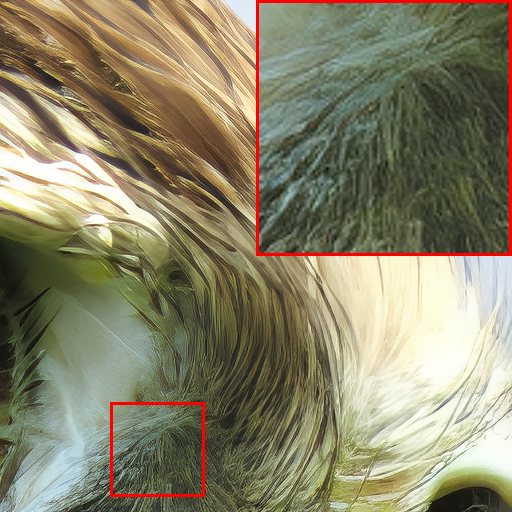}
        \vspace{-1.25cm}
        \begin{center}
        {\hspace{2.25cm}\colorbox{gray}{\color{white}77.07}}
        \end{center}
    \end{minipage}
    \vfill
    \begin{minipage}[t]{0.19\linewidth}
        \includegraphics[width=\linewidth]{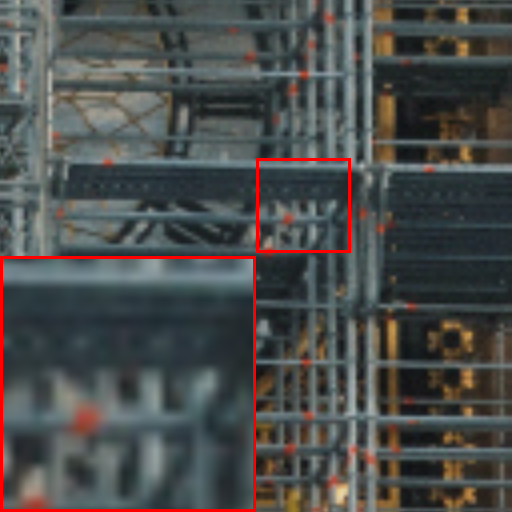}
        \vspace{-1.25cm}
        \begin{center}
        {\hspace{2.25cm}\colorbox{gray}{\color{white}33.52}}
        \end{center}
    \end{minipage}
    \begin{minipage}[t]{0.19\linewidth}
        \includegraphics[width=\linewidth]{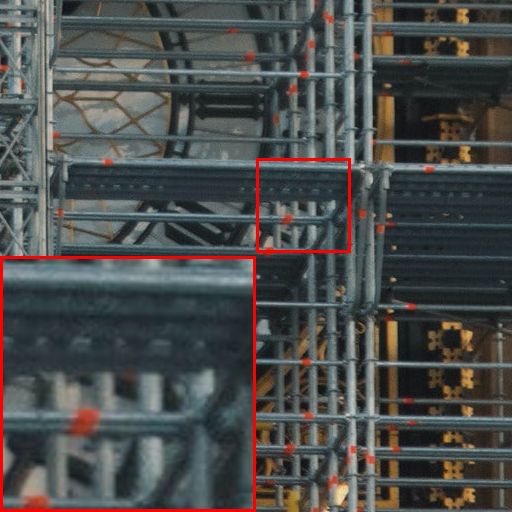}
        \vspace{-1.25cm}
        \begin{center}
        {\hspace{2.25cm}\colorbox{gray}{\color{white}56.01}}
        \end{center}
    \end{minipage}
    \begin{minipage}[t]{0.19\linewidth}
        \includegraphics[width=\linewidth]{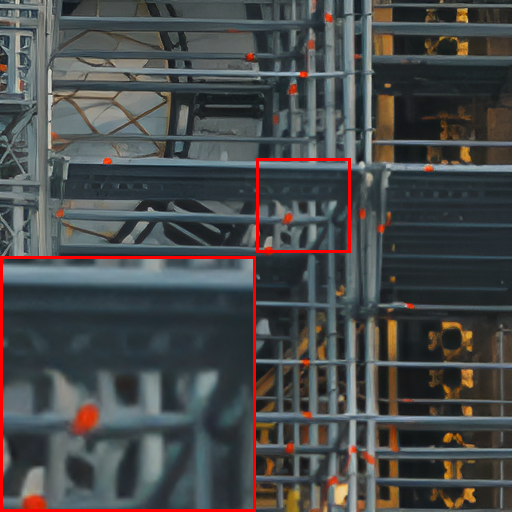}
        \vspace{-1.25cm}
        \begin{center}
        {\hspace{2.25cm}\colorbox{gray}{\color{white}63.01}}
        \end{center}
    \end{minipage}
    \begin{minipage}[t]{0.19\linewidth}
        \includegraphics[width=\linewidth]{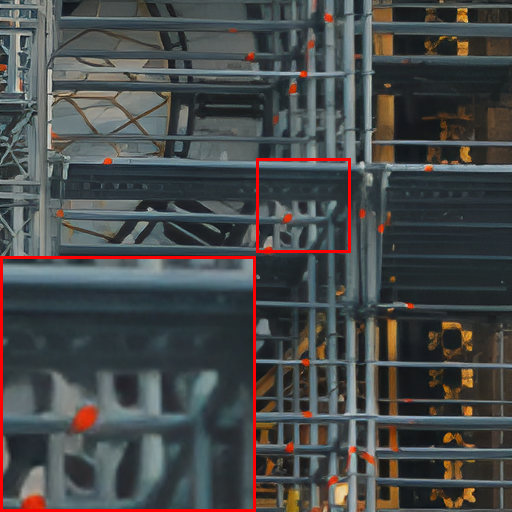}
        \vspace{-1.25cm}
        \begin{center}
        {\hspace{2.25cm}\colorbox{gray}{\color{white}67.50}}
        \end{center}
    \end{minipage}
    \begin{minipage}[t]{0.19\linewidth}
        \includegraphics[width=\linewidth]{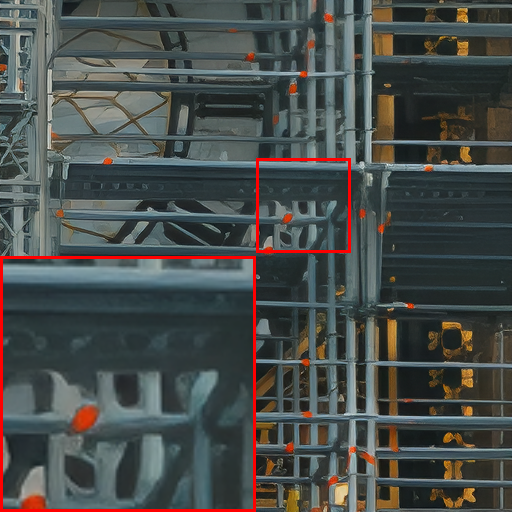}
        \vspace{-1.25cm}
        \begin{center}
        {\hspace{2.25cm}\colorbox{gray}{\color{white}71.15}}
        \end{center}
    \end{minipage}
    \begin{minipage}[t]{0.19\linewidth}
    \centering{LR}
    \end{minipage}%
    \begin{minipage}[t]{0.20\linewidth}
    \centering{Original GT}
    \end{minipage}%
    \begin{minipage}[t]{0.19\linewidth}
    \centering{RESRGAN-UPos}
    \end{minipage}%
    \begin{minipage}[t]{0.20\linewidth}
    \centering{RESRGAN-AMO}
    \end{minipage}%
    \begin{minipage}[t]{0.19\linewidth}
    \centering{RESRGAN-AMO+FT}
    \end{minipage}
    \\
    \caption{
        \textbf{Qualitative results with NR-IQA Guidance.}
        Following the notation of Table~\ref{tab:mainresults},
        columns 3-5 are (top 2 rows) SwinIR-UPos, SwinIR-AMO, and SwinIR-AMO + FT, and (bottom 2 rows) Real-ESRGAN-UPos, Real-ESRGAN-AMO, and Real-ESRGAN-AMO + FT. 
        We show MUSIQ scores in insets.
        Qualitatively, we see improved performance as we move from `UPos' to `AMO' to `AMO-FT', showcasing superiority of each method over the previous one. %
        Zoom in for details. See also Supp.~\S\ref{supp:sec:moreexamples} for additional examples.
    }
    \vspace{-0.3cm}
    \label{fig:qual}
\end{figure*}

\noindent
\textbf{Setup.}
We consider the HGGT dataset under two degradation settings for the super-resolution (SR) problem:
(a) the standard Real-ESRGAN scenario
(two random degradation rounds)
\cite{wang2021real}
and
(b) a simplified setting with a single round, used in the HGGT paper.
We use the ESRGAN (RRDB-based) architecture \cite{wang2018esrgan,wang2021real} for (a) and SwinIR \cite{liang2021swinir} for (b).
Unless otherwise stated, we use the same training settings as HGGT.
Based on our analysis in \S\ref{sec:analysis}, unless noted otherwise, we use MUSIQ \cite{ke2021musiq} as our NR IQA model, $Q$, for both weighted sampling and direct optimization. %

\noindent
\textbf{Evaluation.}
Following HGGT \cite{chen2023human}, we utilize their Test-100 held-out images for evaluation.
We utilize two low-level distortion metrics, PSNR and SSIM \cite{wang2004image}.
We also use three FR models, which act as mid-level visual metrics:
LPIPS \cite{zhang2018image},
LPIPS-ST \cite{ghildyal2022stlpips}, and
DISTS \cite{ding2020iqa}.
LPIPS-ST \cite{ghildyal2022stlpips} is a \textit{shift-tolerant} form of LPIPS \cite{zhang2018image}, improving robustness to small translations imperceptible to humans but highly damaging to  distortion measures.
Finally, we apply four NR-IQA metrics.
Since we are interested in differentiating high-quality images,
we choose MUSIQ \cite{ke2021musiq} and TOPIQ \cite{chen2024topiq} as they have the best positive misalignment scores (see Table~\ref{tab:analysis_HGGT}).
We also include NIMA \cite{talebi2018nima} and Q-Align \cite{wu2024qalign} based on their complementarity with MUSIQ (see \S\ref{analysis:sbs}), which is directly optimized.
Since Test-100 has multiple positive GTs per image, evaluations with reference-based metrics are averaged across all positives.

\noindent
\textbf{Baselines.}
We compare to the SOTA ``positives-only'' model for HGGT, which we denote UPos, as it uses a uniform distribution over positive samples.
We include two human-annotation-free baselines that do not incorporate NR-IQA:
``OrigsOnly'', which trains only with original (non-enhanced) GT,  and ``Rand'', which randomly chooses a supervisory image from among \textit{all} potential GTs.
For our methods, we can choose (a) an IQA-based sampling type and (b) IQA-based fine-tuning settings.
The different sampling methodologies (SMA, SMP, and AMO) are described in \S\ref{sec:methods:rs}, while we denote the use of fine-tuning (see \S\ref{sec:methods:do}) with the ``FT'' moniker.
We also consider two main FT variations, FT$_\text{IG}$ and  FT$_\text{HP}$, described in \S\ref{sec:ablations}.
Our primary method combines the best settings for both IQA-based sampling and optimization: the AMO+FT scheme.

{
\setlength{\tabcolsep}{6pt}
\begin{table*}[t]
    \centering
    \resizebox{\linewidth}{!}{%
    \begin{tabular}{c|c|ccccc|cccc}
    \toprule
        \multirow{2}{*}{Model} & \multirow{2}{*}{\NH} &
            \multicolumn{2}{c}{FR Low-Lev.~Dist.} &
            \multicolumn{3}{c}{FR Mid-Lev.~Dist.} &
            \multicolumn{4}{c}{NR  High-Lev. Perceptual Quality} \\
          & &
            PSNR $\uparrow$ & SSIM $\uparrow$ & 
            LPIPS $\downarrow$  & LPIPS-ST $\downarrow$ & DISTS $\downarrow$ &
            MUSIQ $\uparrow$ & NIMA $\uparrow$ & Q-Align $\uparrow$ & TOPIQ $\uparrow$ \\\midrule \midrule
        Gold Standard & \redcross & -- & --  & -- & -- & -- & 69.64 & 5.28 & 3.78 & 0.69 \\
        \hline
        SwinIR-OrigsOnly & \greencheck &
            \frst 22.72 & \frst 0.652 & 0.227 & 0.174 & 0.162 & 59.47 & 4.87 & 3.17 & 0.48 \\
        SwinIR-Rand & \greencheck &
            \scnd 22.45 & \scnd 0.650 & 0.180 & 0.139 & 0.131 & 65.27 & 5.11 & 3.52 & 0.59 \\    
        SwinIR-UPos$^*$ & \redcross &
            \third 22.30 & \third 0.647 & 0.169 & 0.129 & \scnd 0.123 & 66.39 & 5.16 & 3.56 & 0.62 \\
        SwinIR-SMA & \greencheck  &
            22.27 & 0.646 & 0.171 & 0.129 & \third 0.124 & 66.73 & 5.16 & 3.60 &  0.63 \\
        SwinIR-SMP & \redcross & 
            22.29 & \third 0.647 & 0.171 & 0.130 & \third 0.124 & 66.83 & 5.17 & 3.62 &  0.62 \\
        SwinIR-AMO & \greencheck &
            22.08 & 0.641 & \scnd 0.167 & 0.124 & \scnd 0.123 & 68.08 & 5.21 & 3.67 &  \third 0.66 \\ \hline
        SwinIR-UPos + FT$_{\text{HP}}$ & \redcross  & %
            22.17 & 0.642 & \frst 0.166 & \third 0.123 & \frst 0.122 & 68.38 & 5.23 & 3.64 & 0.65 \\
        SwinIR-UPos + FT$_{\text{IG}}$ & \redcross & %
           22.03 & 0.635 & \third 0.168 & \scnd 0.122 & \scnd 0.123 & \third 69.37 & \third 5.24 & \third 3.69 & \third 0.66 \\
        SwinIR-UPos + FT & \redcross &
            22.01 & 0.633 & 0.169 & \third 0.123 & \third 0.124 & \scnd 69.70 & \scnd 5.26 & \scnd 3.70 & \scnd 0.67 \\
        SwinIR-AMO + FT & \greencheck &
            21.77 & 0.624 & 0.174 & \frst 0.121 & 0.128 & \frst 70.81 & \frst 5.29 & \frst 3.75 & \frst 0.70 \\
        \midrule \midrule
        RESRGAN-OrigsOnly & \greencheck &
            \frst 22.10 & \frst 0.618 & 0.283 & 0.229 & 0.185 & 57.91 & 4.84 & 2.99 & 0.46 \\
        RESRGAN-Rand & \greencheck &
            \scnd 21.66 & \scnd 0.611 & 0.234 & 0.190 & 0.160 & 64.82 & 5.18 & 3.40 & 0.60 \\ 
        RESRGAN-UPos$^*$ & \redcross &
           \third 21.54 & \third 0.608 & 0.233 & 0.192 & \third 0.158 & 65.93 & 5.25 & 3.47 & 0.63 \\
        RESRGAN-SMA & \greencheck & 
           21.46 & 0.606 & \third 0.227 & 0.182 & \scnd 0.157 & 65.87 & 5.23 & 3.46 & 0.63 \\
        RESRGAN-SMP & \redcross & 
           21.44 & 0.607 & \scnd 0.226 & 0.182 & \frst 0.156 & 66.66 & 5.24 & 3.51 & 0.64 \\
        RESRGAN-AMO & \greencheck &
            21.28 & 0.602 & \frst 0.224 & \third 0.178 & \frst 0.156 & 67.86 & 5.29 & 3.56 & 0.66 \\ \hline
        RESRGAN-UPos + FT$_{\text{HP}}$ & \redcross &
           21.30 & 0.595 & \scnd 0.226 & \scnd 0.175 & \third 0.158 & 70.28 & \third 5.32 & 3.65 & \third 0.69 \\
        RESRGAN-UPos + FT$_{\text{IG}}$ & \redcross &
           21.14 & 0.586 & 0.236 & 0.182 & 0.160 & \scnd 72.01 & \scnd 5.35 & \frst 3.70 & \scnd 0.70 \\
        RESRGAN-UPos + FT & \redcross &
           21.09 & 0.580 & 0.235 & 0.179 & 0.163 & \frst 72.69 & \frst 5.37 & \scnd 3.69 & \frst 0.71 \\
        RESRGAN-AMO + FT & \greencheck &
            21.02 & 0.581 & 0.228 & \frst 0.169 & 0.161 & \third 71.67 & \scnd 5.35 & \third 3.68 & \frst 0.71\\ 
            \bottomrule
    \end{tabular}
    }
    \caption{
        \textbf{Evaluation on held-out HGGT Test-100.} 
        ``FR Low-Lev Dist'' refers to full-reference low-level distance metrics; ``FR Mid-Lev Dist'' and ``NR  High-Lev. Perceptual Quality'' 
            refer to full-reference and no-reference perceptual metrics, respectively.
        Second column (\includegraphics[height=0.3cm, trim=0 40mm 0 0]{images/NH.png}) indicates that a method works with \textit{no human GT ranking data} (\greencheck), or requires such GT annotations (\redcross).
        ``Gold Standard'' shows the average of best metric value per quintuplet of test GTs.
        ``OrigsOnly'' means no multimodal supervision (no enhanced GT).
        ``Rand'' signifies random GT choice (from both enhanced and original), 
            which requires no human annotation, while
        ``UPos'' denotes the ``positives-only'' scenario 
            (uniform sampling from human-ranked positives), 
         the SoTA baseline method from HGGT (marked by $^*$).
        ``FT'' refers to fine-tuning (direct optimization);
        ``FT$_{\text{HP}}$''
        denotes using a higher perceptual loss weight ($\lambda_P$), and ``FT$_{\text{IG}}$'' the inclusion of GAN loss.
    }
    \vspace{-0.4cm}
    \label{tab:mainresults}
\end{table*}
}

\subsection{Empirical Results}
\label{sec:empresults}

Our results on HGGT Test-100 are displayed in Table~\ref{tab:mainresults} and Fig.~\ref{fig:qual}. 
See Supp.~\S\ref{supp:sec:realsr} for RealSR results as well.

\noindent
\textbf{Multimodal Training Boosts Performance.}
As in HGGT, we can see the impact of enhanced GTs by comparing OrigsOnly to UPos, which has greatly improved perceptual quality (LPIPS, DIST, and NR-IQA).
We also consider the Rand baseline, showing that even randomly sampling enhanced GTs is helpful for perceptual quality but not sufficient to reach UPos performance, the SoTA method from HGGT, enabled by \textit{human} filtering of low-quality GTs.

\noindent
\textbf{Neural IQA Sampling Outperforms Human Rankings.}
We next investigate whether IQA-based sampling (SMA, SMP, and AMO) can outperform UPos, which relies on human rankings.
On SwinIR, LPIPS and DISTS remain largely unchanged, but LPIPS-ST and the NR metrics show small improvements, especially for AMO.
On Real-ESRGAN, sampling with IQA improves both LPIPS and LPIPS-ST, but only AMO shows substantial improvements on the NR metrics.
Surprisingly, despite access to human labels, 
SMP is very similar to SMA,
maintaining nearly identical performance on low and mid level distortion measures, with a marginal boost in NR image quality.
In general, we find AMO is consistently superior to both SMA and SMP, which suggests that selecting for quality (especially at the fine-grained level demanded in the online setting) is more important than simply having multiple GTs; AMO is also measurably better than UPos in terms of perceptual quality, despite the lack of access to human annotation.

\noindent
\textbf{IQA Fine-tuning Improves both Human and Neural Sampling.}
We examine the impact of fine-tuning (FT) on NR-IQA.
When used on top of human data, denoted UPos+FT, LPIPS and DISTS are the same (SwinIR) or slightly worse (RealESRGAN), but NR-IQA metrics uniformly improve, as well as, interestingly, LPIPS-ST.
This trade-off of mid-level perceptual distortion for high-level quality is effectively an extension of the previously observed balance, between pixel-level distortion versus perceptual metrics.
Indeed, we still see the latter compromise here, in that FT always damages low-level distortion metrics (PSNR and SSIM), despite the increases in perceptual quality. This is expected, since NR-based FT does not optimize to a particular SR solution, let alone the one(s) in the dataset.
In addition, the results with LPIPS-ST indicate that it is more perceptual than LPIPS or DIST (i.e., more NR-IQA-like, though still full-reference). 
Overall, the performance boosts with FT suggest that useful information can be extracted from neural NR-IQA models, 
even on top of a model with access to human annotations, 
providing a simple mechanism for improving image quality in SR models.

\noindent
\textbf{Upper-bounding NR-IQA Evaluation Performance.}
In addition, we compute ``gold standard'' NR-IQA values for the GT Test data, taking the best score among the original and enhanced images, providing a soft upper-bound on NR scores, if one were able to exactly reproduce the ``best'' GT via the SR network.
We find that altered sampling generally does not reach these values, but FT is able to reach and even surpass them in several scenarios.

\noindent
\textbf{Superior Perceptual Quality via Neural Sampling and Fine-Tuning.}
Finally, we test the natural unification of IQA-based neural sampling with IQA-driven FT, denoted by the AMO+FT setting.
We find that this combination surpasses the SoTA HGGT approach (UPos), in terms of perceptual quality, despite \textit{not using any human annotations}.
Specifically, for SwinIR, AMO+FT incurs a small penalty ($\sim$3-4\%) on LPIPS and DISTS, but improves LPIPS-ST by $\sim$6\% and is superior to every other method according to NR metrics.
In the RealESRGAN setting, compared to UPos, AMO+FT again obtains a large improvement on LPIPS-ST ($\sim$12\%), with neglible changes on the other mid-level metrics (${\sim}$2\%).
It also soundly surpasses UPos according to \textit{every} NR metric.
Interestingly, RealESRGAN-AMO+FT does not outperform RealESRGAN-UPos+FT; however, note that the latter has access to human annotations.
We also present a user study in Supp.~\S\ref{supp:sec:userstudy}, finding a preference for AMO+FT over UPos.  
Altogether, these results suggest 
(i) human annotated rankings on multiple GTs, at least when used naively for SR training, can be easily surpassed via neural NR-IQA scores, and
(ii) considerable improvements to the perceptual quality of SR models can be attained through automated means, by simply applying existing NR-IQA models.

\subsection{Ablations and Variations}
\label{sec:ablations}

\noindent
\textbf{IQA Optimization Refines the Perception--Distortion Trade-off.}
As noted, our results show a trade-off between mid-level perceptual and NR-IQA metrics, reminiscent of the classic perception--distortion curve \cite{blau2018perception,blau2019rethinking} (which we also observe, via PSNR and SSIM).
We can control this mid-versus-high-level perceptual tradeoff, by simply changing the FT loss weights.
For instance, comparing UPos+FT to UPos+FT$_\text{HP}$ (which increases $\lambda_P$), we see that mid-level metrics all improve, while all NR metrics decline.

\noindent
\textbf{Discriminators as IQA.}
One can naturally interpret the discriminator (or critic) of a GAN as a form of NR-IQA model -- in fact, it is one specialized to the errors and artifacts of the SR function we are training.
However, the FT$_\text{IG}$ scenario, which keeps the GAN loss during FT, does not greatly impact results (compared to the FT setting); in fact, it largely induces very slight declines.
We also tried UPos+FT without an NR-IQA model, instead simply upweighting the GAN loss (treating the critic as an IQA model); however, this results in uniformly worse NR scores, with little change to mid-level metrics
(see Supp.\ \S\ref{supp:ablations}).

\noindent
\textbf{Alternative NR-IQA.}
While we selected MUSIQ based on our analysis (\S\ref{sec:analysis}), we also tested FT with an alternative IQA model, PaQ-2-PiQ, based on its high score in Table~\ref{tab:analysis_phaseII} 
(see Supp.\ \S\ref{supp:ablations}).
We find that NIMA, Q-Align, TOPIQ, and (unsurprisingly) MUSIQ all decline, for both SwinIR and RealESRGAN.
Nevertheless, it is plausible that a different IQA model (or combination thereof), particularly if fine-tuned for SR, would provide a superior learning signal.

\section{Conclusion}
\label{sec:conc}
As an ill-posed inverse problem, SR struggles with the dichotomy between perceptual quality and reference fidelity. 
Prior research utilized multiple GTs and human annotations to mitigate  this trade-off.
In contrast, herein, we focus on improving perceptual quality via neural IQA, enabling us to eschew human annotations. 
We first %
analyze existing NR-IQA methods, discerning a candidate for adoption in training, as well as complementary models for evaluation.
Then, we devised two ways to apply NR-IQA to SR training:
(i) IQA-weighted multimodal GT sampling and
(ii) regularized optimization of NR quality.
When jointly utilized, our approach outperforms the existing SoTA, which relies on human data, in terms of NR metrics, without sacrificing mid-level reference-based quality scores. 
We hope it enables future investigation into NR-IQA for SR, the connection between generative modelling and IQA, and SR with domain shift, as IQA does not need paired GT.

\newif\ifSUPPONLY%
\SUPPONLYfalse%
\clearpage\newpage


\clearpage
\maketitlesupplementary

\begin{table*}[t]
    \centering
    \resizebox{\textwidth}{!}{%
    \begin{tabular}{l c | l c | l c | l c | l c}
        \toprule
        Method & Acc (\%) & Method & Acc (\%) & Method & Acc (\%) & Method & Acc (\%) & Method & Acc (\%) \\
        \midrule
        paq2piq	&	76.41	&	arniqa-kadid	&	71.48	&	tres	&	69.98	&	arniqa-clive	&	66.81	&	brisque\_matlab	&	61.00	\\
        nima	&	74.91	&	arniqa-flive	&	71.30	&	clipiqa+\_vitL14\_512	&	69.98	&	arniqa-spaq	&	66.73	&	wadiqam\_nr	&	60.30	\\
        musiq	&	74.47	&	topiq\_nr-spaq	&	71.30	&	musiq-paq2piq	&	69.98	&	arniqa	&	66.46	&	topiq\_nr-flive	&	58.19	\\
        liqe	&	74.03	&	arniqa-csiq	&	71.21	&	maniqa-pipal	&	69.63	&	musiq-ava	&	66.37	&	ilniqe	&	57.92	\\
        arniqa-tid	&	74.03	&	musiq-spaq	&	70.86	&	clipiqa+\_rn50\_512	&	69.10	&	nrqm	&	65.05	&	niqe	&	56.43	\\
        qalign	&	73.77	&	nima-vgg16-ava	&	70.77	&	dbcnn	&	68.49	&	cnniqa	&	63.73	&	brisque	&	55.11	\\
        topiq\_nr	&	73.06	&	maniqa	&	70.51	&	clipiqa	&	68.40	&	tres-flive	&	63.29	&	niqe\_matlab	&	51.94	\\
        hyperiqa	&	72.27	&	clipiqa+	&	70.25	&	arniqa-live	&	68.05	&	pi	&	62.41	&	piqe	&	46.21	\\
        liqe\_mix	&	71.48	&	maniqa-kadid	&	70.16	&		&		&		&		&		&		\\
        \bottomrule
    \end{tabular}
    }
    \caption{\textbf{Phase I analysis on SBS180K dataset.} Accuracy of 20 NR-IQA metrics and their variants on the subset (1212 image pairs) of train set of SBS180K dataset. We denote a metric by its `Model Name' as defined in \texttt{IQA-PyTorch} toolbox (\url{https://iqa-pytorch.readthedocs.io/en/latest/ModelCard.html}). We use the default configuration for all metrics and their variants.}
    \label{tab:Phase I_supp}
\end{table*}

\section{Complete Analysis of NR-IQA metrics}
In \S \ref{analysis:sbs} of the main paper, we present accuracy of only top 7 NR-IQA metrics on the subset of SBS180K~\cite{khrulkov2021neural} train set. Here, in Table~\ref{tab:Phase I_supp}, we present accuracy of 20 NR-IQA metrics and their variants (42 in total) on the same subset. We make two main observations. First, unsurprisingly, recent NR-IQA metrics (e.g. PaQ-2-PiQ~\cite{ying2020patches}, MUSIQ~\cite{ke2021musiq}, Q-Align~\cite{wu2024qalign}) are more aligned with human preferences than the classical ones (e.g. NIQE~\cite{zhang2015feature} and BRISQUE~\cite{mittal2012no}), calling for wider adaptation of more recent metrics in evaluating SR models. Second, the IQA dataset on which the metric is trained affects its accuracy in determining human preference for SR. For instance, TOPIQ~\cite{chen2024topiq} trained using KonIQ~\cite{hosu2020koniq} dataset is more aligned with human judgement (73.06\%) than the one trained using FLIVE~\cite{ying2020patches} dataset (58.19\%). Results indicate an opportunity to create a no-reference IQA dataset exclusively for training NR metrics for SR. 

\subsection{Remark on NR-IQA Choices}
\label{supp:sec:nriqachoices}
As discussed in \S\ref{sec:analysis}, our choice of MUSIQ for weighted sampling and fine-tuning comes from several considerations. First, on SBS-180K, MUSIQ is highly performant (see Phase II analysis of \S\ref{analysis:sbs}). Second, on HGGT (\S\ref{analysis:hggt}), MUSIQ has the best positive misalignment, meaning it is the least likely to misrank a {positive}. It is also relatively efficient for both inference and back-propagation.

In our application, distinguishing between the quality of positives would seem to be more significant, since we are ideally training the SR model in a manner that focuses on the highest quality images (i.e., incorrect ordering of the lower-ranked images will not affect our method; hence, fine-grained differentiation between the high-ranked images is more important).
Nevertheless, it is true that MUSIQ (like all NR-IQA models evaluated here) does not perform well on negative misalignment. 
However, we do not expect this to have a large impact on training, due to the \textit{rarity} of negatives. 
Specifically, in HGGT-train, only ${\sim}$6\% of tuples contain negatives and, among those, MUSIQ ranks a negative the highest in ${\sim}$34\% of cases. Thus, our AMO model will be exposed to a negative in only ${\sim}$2\% of examples. Hence, merely for numerical magnitude, discernment for positives is likely to be more impactful than for negatives. Of course, this reasoning is somewhat specific to the HGGT setup.

Further, in terms of evaluation, note that we choose NIMA and Q-Align specifically because they perform best on the SBS-180K samples on which MUSIQ fails. Ideally, this complementarity would help ensure that errors induced by shortcomings of MUSIQ could potentially be detected by the other NR-IQA metrics. Nevertheless, as seen in Table~\ref{tab:suppresults}, our experiments with PaQ-2-PiQ (which was among the best models according to Table~\ref{tab:analysis_phaseII}) show MUSIQ outperforms it.

Regardless, our method does not specifically require the use of MUSIQ. Indeed, we believe further advancements in NR-IQA models (e.g., approaches specific to SR image quality, adversarially robust models) will be applicable to our method as well.

{
\setlength{\tabcolsep}{4pt}
\begin{table*}[t]
    \centering
    \resizebox{\linewidth}{!}{%
    \begin{tabular}{c|ccc|ccccc|cccc}
    \toprule
        \multirow{2}{*}{Model} & 
            \multirow{2}{*}{NR} &
            \multirow{2}{*}{$\lambda_A$} &
            \multirow{2}{*}{\NH} &
            \multicolumn{2}{c}{FR Low-Lev.~Dist.} &
            \multicolumn{3}{c}{FR Mid-Lev.~Dist.} &
            \multicolumn{4}{c}{NR  High-Lev. Perceptual Quality} \\
          & & & &
            PSNR $\uparrow$ & SSIM $\uparrow$ & 
            LPIPS $\downarrow$  & LPIPS-ST $\downarrow$ & DISTS $\downarrow$ &
            MUSIQ $\uparrow$ & NIMA $\uparrow$ & Q-Align $\uparrow$ & TOPIQ $\uparrow$ \\\midrule \midrule
        Gold Standard & -- & -- & \redcross & -- & --  & -- & -- & -- & 69.64 & 5.28 & 3.78 & 0.69 \\
        \hline
        SwinIR-UPos$^*$  & -- & -- & \redcross &
            \frst 22.30 & \scnd 0.647 & \third 0.169 & 0.129 & \third 0.123 & 66.39 & 5.16 & 3.56 & 0.62 \\
        SwinIR-UPos + FT$_{\text{HP}}$ & M & 0 &  \redcross & %
            22.17 & 0.642 & \frst 0.166 &  0.123 & \scnd 0.122 & 68.38 & 5.23 & 3.64 & 0.65 \\
        SwinIR-UPos + FT$_{\text{IG}}$ & M & 0.1 &  \redcross & %
           22.03 & 0.635 & \scnd 0.168 & \third 0.122 & \third 0.123 & \third 69.37 & \third 5.24 &  \third 3.69 & \third 0.66 \\
        SwinIR-UPos + FT$_{\text{NNR,IG}\times 2}$ & -- & 0.2 &  \redcross & %
           \third 22.25 & \third 0.646 & 0.171 & 0.130 & 0.124 & 66.61 & 5.16 & 3.56 & 0.61 \\
        SwinIR-UPos + FT$_{\text{NNR,IG}\times 5}$ & -- & 0.5 & \redcross & %
           22.20 & 0.644 & 0.174 & 0.134 & 0.125 & 66.61 & 5.16 & 3.56 & 0.61 \\
        SwinIR-UPos + FT$_{\text{PaQ2PiQ}}$ & P & 0 &  \redcross & %
           \scnd 22.29 & \frst 0.649 & \frst 0.166 & \frst 0.120 & \frst 0.121 & 67.29 & 5.18 & 3.58 & 0.62 \\
        SwinIR-UPos + FT & M & 0 &  \redcross &
            22.01 & 0.633 & \third 0.169 &  0.123 &  0.124 & \scnd 69.70 & \scnd 5.26 & \scnd 3.70 & \scnd 0.67 \\
        SwinIR-AMO + FT & M & 0 & \greencheck &
            21.77 & 0.624 & 0.174 & \scnd 0.121 & 0.128 & \frst 70.81 & \frst 5.29 & \frst 3.75 & \frst 0.70 \\
        \midrule \midrule
        RESRGAN-UPos$^*$ & -- & -- & \redcross &
           \frst 21.54 & \frst 0.608 & \third 0.233 & 0.192 & \scnd 0.158 & 65.93 & 5.25 & 3.47 & 0.63 \\
        RESRGAN-UPos + FT$_{\text{HP}}$ & M & 0 &  \redcross &
           21.30 & 0.595 & \frst 0.226 & \scnd 0.175 & \scnd 0.158 & 70.28 & \third 5.32 & 3.65 & \third 0.69 \\
        RESRGAN-UPos + FT$_{\text{IG}}$ & M & 0.1 &  \redcross &
           21.14 & 0.586 & 0.236 & 0.182 & \third 0.160 & \scnd 72.01 & \scnd 5.35 & \frst 3.70 & \scnd 0.70 \\
        RESRGAN-UPos + FT$_{\text{NNR,IG}\times 2}$ & -- & 0.2 &  \redcross & %
           \third 21.35 & \third 0.600 & 0.234 & 0.191 & \frst 0.157 & 65.94 & 5.22 & 3.45 & 0.63 \\
        RESRGAN-UPos + FT$_{\text{NNR,IG}\times 5}$ & -- & 0.5 &  \redcross & %
           21.25 & 0.598 & 0.237 & 0.195 & \scnd 0.158 & 65.78 & 5.22 & 3.46 & 0.63 \\
        RESRGAN-UPos + FT$_{\text{PaQ2PiQ}}$ & P & 0 &  \redcross & %
           \scnd 21.46 & \scnd 0.605 & \scnd 0.228 & 0.182 & \frst 0.157 & 67.26 & 5.22 & 3.51 & 0.64 \\   
        RESRGAN-UPos + FT & M & 0 &  \redcross &
           21.09 & 0.580 & 0.235 & \third 0.179 & 0.163 & \frst 72.69 & \frst 5.37 & \scnd 3.69 & \frst 0.71 \\
        RESRGAN-AMO + FT & M & 0 &  \greencheck &
            21.02 & 0.581 & \scnd 0.228 & \frst 0.169 & 0.161 & \third 71.67 & \scnd 5.35 & \third 3.68 & \frst 0.71\\ 
            \bottomrule
    \end{tabular}
    }
    \caption{
        \textbf{Additional evaluation on held-out HGGT Test-100.} 
        As in Table \ref{tab:mainresults} in the main paper,
            ``FR Low-Lev Dist'' refers to full-reference low-level distance metrics; ``FR Mid-Lev Dist'' and ``NR  High-Lev. Perceptual Quality'' 
            refer to full-reference and no-reference perceptual metrics, respectively.
        Second column (\includegraphics[height=0.3cm, trim=0 40mm 0 0]{images/NH.png}) indicates that a method works with \textit{no human GT ranking data} (\greencheck), or requires such GT annotations (\redcross).
        ``Gold Standard'' shows the average of best metric value per quintuplet of test GTs.
        ``UPos'' denotes the ``positives-only'' scenario (uniform sampling from human-ranked positives), 
             the SoTA baseline method from HGGT (marked by $^*$).
        ``FT'' refers to fine-tuning (direct optimization):
        ``FT$_{\text{IG}}$'' includes the adversarial loss during FT,
        ``FT$_{\text{NNR,IG}\times 2}$'' and
        ``FT$_{\text{NNR,IG}\times 5}$'' have no NR term during FT, but increase the GAN loss (two and five times, respectively), and finally
        ``FT$_{\text{PaQ2PiQ}}$'' replaces MUSIQ with PaQ-2-PiQ.
        The NR column denotes which NR-IQA model is used (M: MUSIQ, P: PaQ-2-PiQ, --: None),
            while $\lambda_A$ is the adversarial loss weight 
            (the standard HGGT default for training is 0.1).
        We also show our best method: AMO+FT, which combines IQA-based sampling with our standard FT settings, for comparison.
        Note that AMO+FT is the only method here that \textit{does not use human annotations}. 
        We remark also that the NR-IQA models have the following ranges:
        MUSIQ (0-100), NIMA (0-10), Q-Align (1,5), and TOPIQ (0-1).
    }
    \vspace{-0.4cm}
    \label{tab:suppresults}
\end{table*}
}

\begin{figure*}[t]
    \centering
    \includegraphics[width=0.999\textwidth]{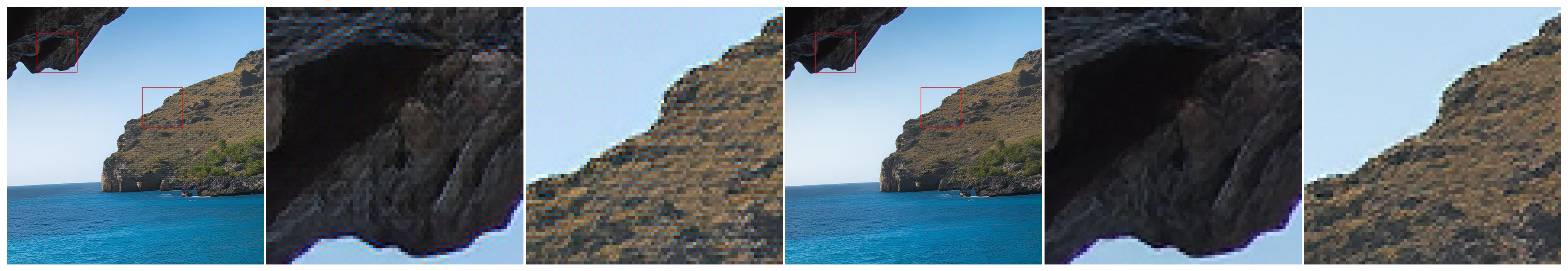}
    \caption{
        \textbf{Structured Noise due to naive NR-IQA optimization.}
        The left three insets show an image and two close-ups that was fine-tuned \textit{without} LoRA, whereas the right three show the effect of using LoRA.
        Note the patterns that form in the sky and the strangely coloured pixels that appear around certain edges (e.g., the blue/red grid in the second inset) when LoRA is not used. 
    }
    \label{fig:supp:advartifacts}
\end{figure*}

\section{Methodological Details}
\label{supp:methods}

\subsection{Sampling Details}
The altered sampling (\S\ref{sec:methods:rs}) is trained identically to the standard HGGT version, just replacing the uniform nature of the GT sampling.
The only additional parameter is the temperature, $\tau$, which we set to 10 for both SMA and SMP.

\subsection{Hardware and Timing}
Similar to HGGT, we train on four A100 GPUs for 300K iterations.
This takes ${\sim}23$ and ${\sim}32$ hours for SwinIR and RealESRGAN, respectively, with an additional ${\sim}3.5$ hours for fine-tuning.

\subsection{Fine-Tuning Details}

Unless otherwise noted, we use the same training parameters as HGGT. %
We fine-tune for only 20,000 steps and set $\lambda_Q = 0.05$ as the FT MUSIQ weight.
For SwinIR and RealESRGAN, respectively, we change the learning rate to $5\times 10^{-6}$ (halved at 5K steps) and $5 \times 10^{-5}$.
Recall that, by default, the adversarial loss is not used (i.e., $\lambda_A = 0$) during FT (but see \S\ref{supp:ablations}).
Architecturally, LoRA weights are inserted slightly differently:
on SwinIR \cite{liang2021swinir},
only the multilayer perceptions are altered (rank 48),
while on the convolutional RealESRGAN \cite{wang2018esrgan,wang2021real},
only the layers in the Residual-in-Residual Dense Blocks (RRDBs) are altered (rank 24). %
This follows other works, including the original LoRA paper \cite{hu2021lora}, which only apply LoRA-based fine-tuning to a subset of layers
(e.g., see \cite{galim2024parameter,kwon2023datainf,lin2025tracking}).
Recall that LoRA cannot increase the capacity (i.e., expressive capability) of the networks (as the new weights can simply be merged into the old ones at inference time, which also prevents any run-time penalty to inference), so comparisons to non-FT models are fair.
Fine-tuning is run for 20K steps, as opposed to the 300K in stage two training.
Brief exploration of hyper-parameters (beyond those considered in \S\ref{sec:results}) yielded minimal changes, likely due to rapid convergence of the low-rank (i.e., low capacity) weights $\phi$. 

\section{Detailed Results on Ablations and Variations}
\label{supp:ablations}

In this section, we consider additional FT variations: 
(i) using a GAN discriminator instead of an NR-IQA model
(using two different loss weights)
and 
(ii) replacing $Q$ (set as MUSIQ) with a different NR-IQA model (PaQ-2-PiQ).
The point of (i) is to check whether the GAN critic, which is effectively an NR-IQA model that has been specialized to the SR model in question, can be used for fine-tuning, instead of a separate NR-IQA model.
For (ii), we wish to check if our choice of optimized NR metric, MUSIQ, is reasonable.

Our results on these variations are in Table \ref{tab:suppresults}.
Since FT optimizes MUSIQ, we focus on the other NR metrics, especially Q-Align and NIMA (since they perform the best on examples where MUSIQ fails; see \S\ref{analysis:sbs}).
First, we find that including the GAN loss in the standard scenario has a slight negative effect on the NR metrics; however, removing the NR metric term and strengthening the adversarial term (i.e., ``FT$_{\text{NNR,IG}\times 2}$'' and ``FT$_{\text{NNR,IG}\times 5}$'') has a significantly more negative impact on the NR evaluations.
This suggests that the critic network \textit{cannot} replace the NR-IQA model, even though it is intuitively similar to one (in that it evaluates the image quality of a single input, which can be used as a learning signal).
We conjecture this is because the critic is trained to detect the idiosyncrasies of its associated generator (at a specific point in time), rather than match human quality estimates; hence, optimizing it more aggressively may reduce those specific issues that the critic has detected, but not necessarily increase general quality.

Second, we tried to replace MUSIQ with PaQ-to-PiQ.
We find that this tends to improve low and mid level distortion (though the relation is less clear for RealESRGAN, especially with LPIPS-ST), but worsens NIMA and Q-Align. We therefore choose to stay with MUSIQ for our main results.
In general, we do not wish to claim that MUSIQ is an optimal starting point for FT; however, it does suggest our analysis is a useful approach to initially identifying a good NR-IQA network. Nevertheless, we suspect that using an alternative NR-IQA model (with sufficient hyper-parameter exploration), fine-tuning a new model, combining multiple models, or training a model specific to SR could all be potentially useful future approaches to improving results.

{
\setlength{\tabcolsep}{4pt}
\begin{table*}[t]
    \centering
    \resizebox{0.999\linewidth}{!}{%
    \begin{tabular}{c|c|ccccc|cccc}
    \toprule
        \multirow{2}{*}{Model} & 
        \multirow{2}{*}{\NH} &
        \multicolumn{2}{c}{FR Low-Lev.~Dist.} &
        \multicolumn{3}{c}{FR Mid-Lev.~Dist.} &
        \multicolumn{4}{c}{NR  High-Lev. Perceptual Quality} \\
         & &
            PSNR $\uparrow$ & SSIM $\uparrow$ & 
            LPIPS $\downarrow$  & LPIPS-ST $\downarrow$ & DISTS $\downarrow$ &
            MUSIQ $\uparrow$ & NIMA $\uparrow$ & Q-Align $\uparrow$ & TOPIQ $\uparrow$ \\\midrule \midrule

        SwinIR-OrigsOnly  & \greencheck &
            \textbf{26.05} & 0.746 & 0.37 & 0.38 & 0.20 & 31.37 & 4.24 & 2.88 & 0.23 \\
        SwinIR-UPos$^*$  & \redcross &
            26.02 & \textbf{0.747} & 0.35 & 0.37 & 0.20 & 33.69 & 4.31 & 2.95 & 0.24 \\
        SwinIR-AMO & \greencheck & %
            25.99 & \textbf{0.747} & 0.34 & 0.37 & \textbf{0.19} & 34.86 & 4.32 & 2.96 & 0.25 \\
        SwinIR-AMO + FT & \greencheck &
            25.96 & 0.742 & \textbf{0.33} & \textbf{0.35} & \textbf{0.19} & \textbf{39.25} & \textbf{4.37} & \textbf{2.99} & \textbf{0.30} \\
        \midrule \midrule

        RESRGAN-OrigsOnly & \greencheck &
            \textbf{25.90} & \textbf{0.758} & \textbf{0.27} & 0.27 & \textbf{0.16} & 46.11 & 4.80 & 3.40 & 0.32 \\
        RESRGAN-UPos$^*$ & \redcross &
            25.45 & 0.750 & 0.28 & 0.26 & 0.17 & 52.74 & 4.95 & 3.53 & 0.41 \\
        RESRGAN-AMO & \greencheck &
            25.22 & 0.745 & 0.28 & 0.25 & 0.17 & 54.73 & 4.97 & 3.57 & 0.45 \\
        RESRGAN-AMO + FT & \greencheck &
            24.71 & 0.718 & 0.32 & \textbf{0.24} & 0.19 & \textbf{65.12} & \textbf{5.03} & \textbf{3.77} & \textbf{0.63}\\ 
            \bottomrule
    \end{tabular}
    }
    \caption{
        \textbf{Additional evaluation on the RealSRv3 \cite{cai2019toward}.} 
        Following Table \ref{tab:suppresults}, we evaluate the four main models on the RealSR V3 dataset, which consists of 100 test images captured using two DSLR cameras (Canon 5D3 and Nikon D810). Our methods (``AMO'' and ``AMO + FT'') achieve the highest no-reference perceptual metric (i.e., NR-IQA) scores, outperforming both ``OrigsOnly'' (without enhanced GT) and ``UPos'' (the SOTA baseline from HGGT, marked by $^*$).
    }
    \vspace{-0.4cm}
    \label{tab:resultsr}
\end{table*}
}

\section{Additional Qualitative Examples}
\subsection{Additional Comparative Samples}
\label{supp:sec:moreexamples}
\begin{figure*}
    \centering
    \begin{minipage}[t]{0.19\linewidth}
        \includegraphics[width=\linewidth]{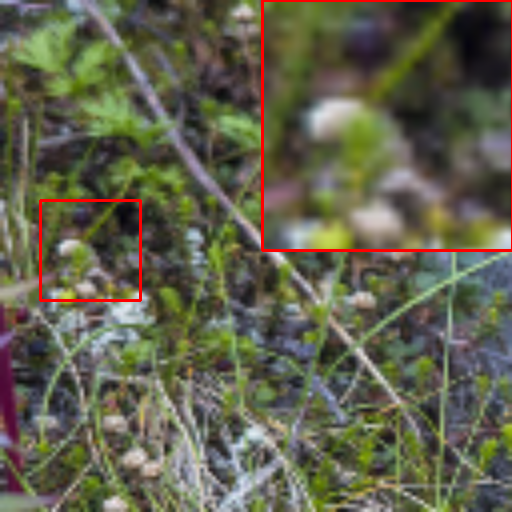}
        \vspace{-1.25cm}
        \begin{center}
        {\hspace{2.25cm}\colorbox{gray}{\color{white}32.77}}
        \end{center}
    \end{minipage}
    \begin{minipage}[t]{0.19\linewidth}
        \includegraphics[width=\linewidth]{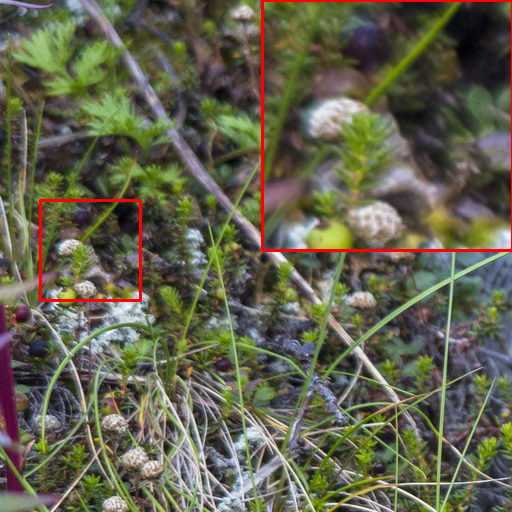}
        \vspace{-1.25cm}
        \begin{center}
        {\hspace{2.25cm}\colorbox{gray}{\color{white}65.83}}
        \end{center}
    \end{minipage}
    \begin{minipage}[t]{0.19\linewidth}
        \includegraphics[width=\linewidth]{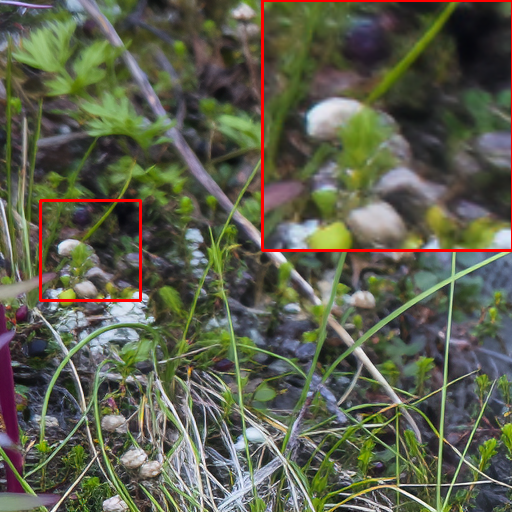}
        \vspace{-1.25cm}
        \begin{center}
        {\hspace{2.25cm}\colorbox{gray}{\color{white}70.61}}
        \end{center}
    \end{minipage}
    \begin{minipage}[t]{0.19\linewidth}
        \includegraphics[width=\linewidth]{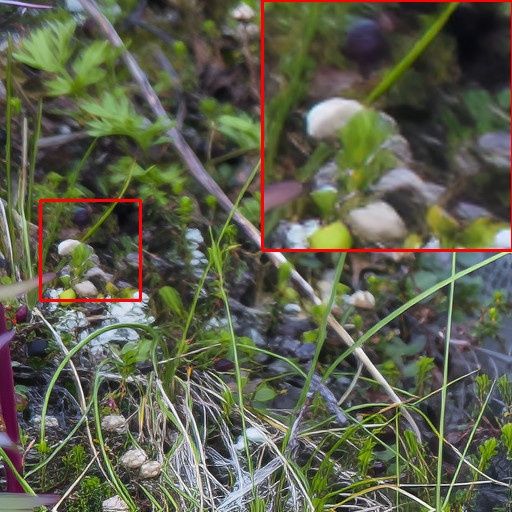}
        \vspace{-1.25cm}
        \begin{center}
        {\hspace{2.25cm}\colorbox{gray}{\color{white}73.07}}
        \end{center}
    \end{minipage}
    \begin{minipage}[t]{0.19\linewidth}
        \includegraphics[width=\linewidth]{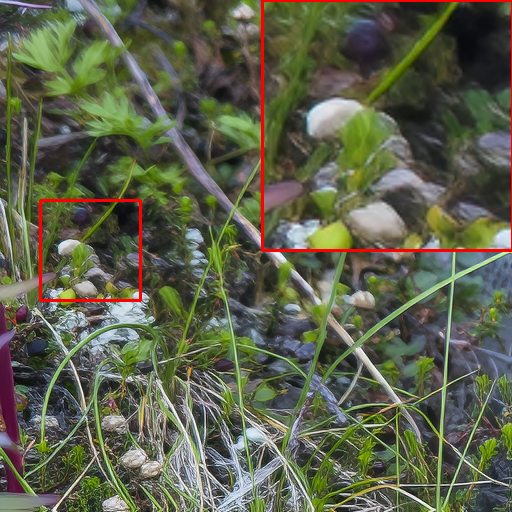}
        \vspace{-1.25cm}
        \begin{center}
        {\hspace{2.25cm}\colorbox{gray}{\color{white}74.91}}
        \end{center}
    \end{minipage}
    
    \vfill
    \begin{minipage}[t]{0.19\linewidth}
        \includegraphics[width=\linewidth]{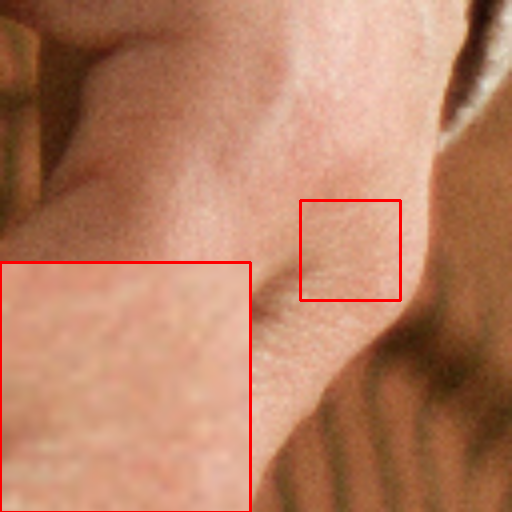}
        \vspace{-1.25cm}
        \begin{center}
        {\hspace{2.25cm}\colorbox{gray}{\color{white}26.45}}
        \end{center}
    \end{minipage}
    \begin{minipage}[t]{0.19\linewidth}
        \includegraphics[width=\linewidth]{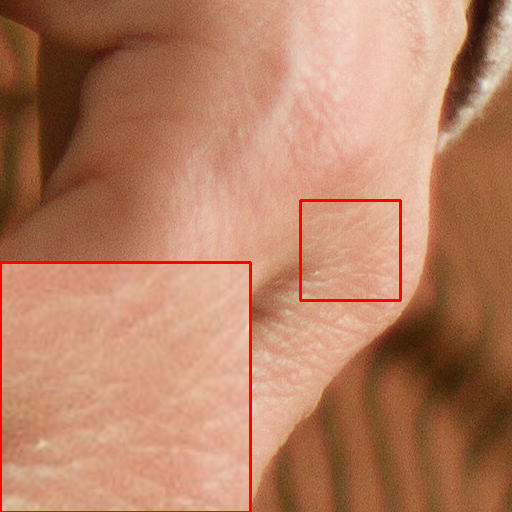}
        \vspace{-1.25cm}
        \begin{center}
        {\hspace{2.25cm}\colorbox{gray}{\color{white}45.26}}
        \end{center}
    \end{minipage}
    \begin{minipage}[t]{0.19\linewidth}
        \includegraphics[width=\linewidth]{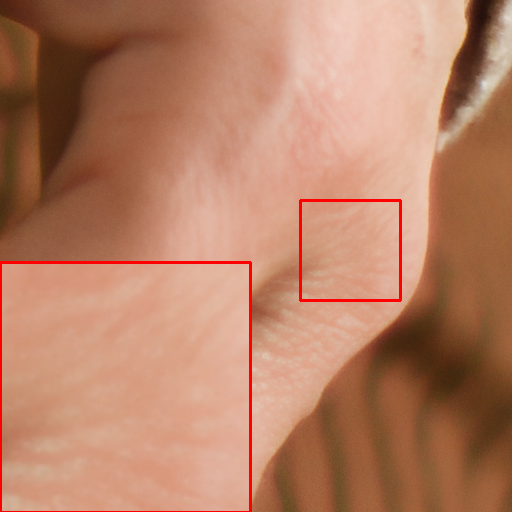}
        \vspace{-1.25cm}
        \begin{center}
        {\hspace{2.25cm}\colorbox{gray}{\color{white}45.60}}
        \end{center}
    \end{minipage}
    \begin{minipage}[t]{0.19\linewidth}
        \includegraphics[width=\linewidth]{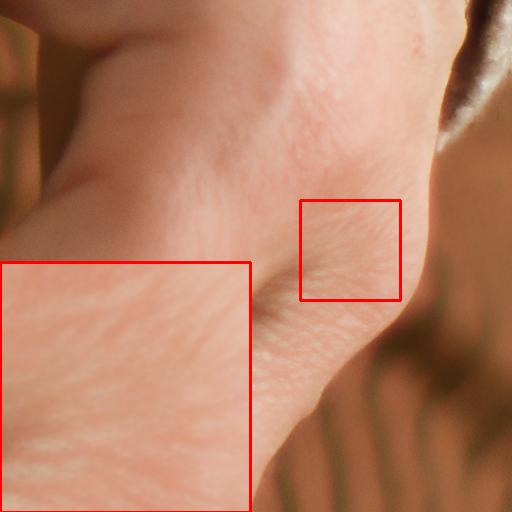}
        \vspace{-1.25cm}
        \begin{center}
        {\hspace{2.25cm}\colorbox{gray}{\color{white}46.71}}
        \end{center}
    \end{minipage}
    \begin{minipage}[t]{0.19\linewidth}
        \includegraphics[width=\linewidth]{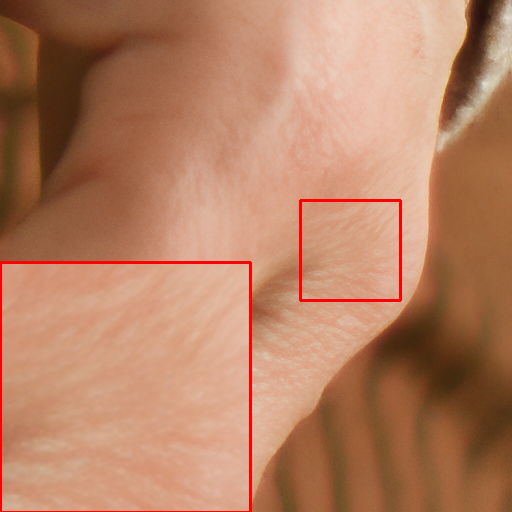}
        \vspace{-1.25cm}
        \begin{center}
        {\hspace{2.25cm}\colorbox{gray}{\color{white}53.20}}
        \end{center}
    \end{minipage}

    \vfill
    \begin{minipage}[t]{0.19\linewidth}
        \includegraphics[width=\linewidth]{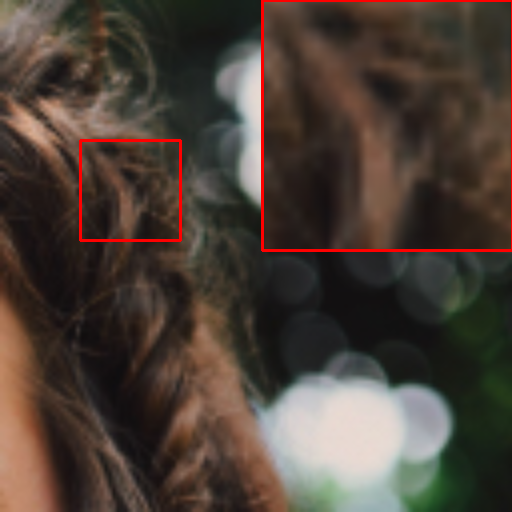}
        \vspace{-1.25cm}
        \begin{center}
        {\hspace{2.25cm}\colorbox{gray}{\color{white}30.75}}
        \end{center}
    \end{minipage}
    \begin{minipage}[t]{0.19\linewidth}
        \includegraphics[width=\linewidth]{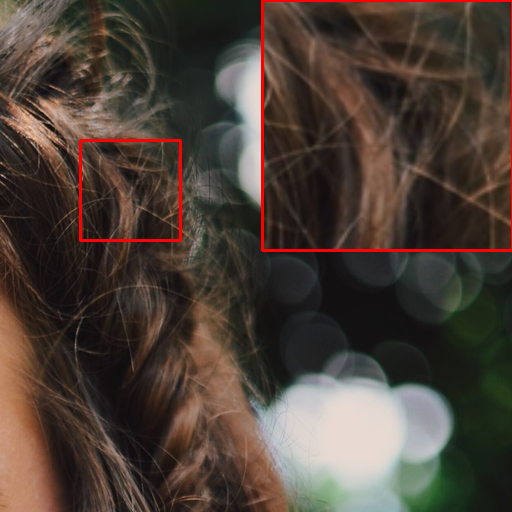}
        \vspace{-1.25cm}
        \begin{center}
        {\hspace{2.25cm}\colorbox{gray}{\color{white}48.81}}
        \end{center}
    \end{minipage}
    \begin{minipage}[t]{0.19\linewidth}
        \includegraphics[width=\linewidth]{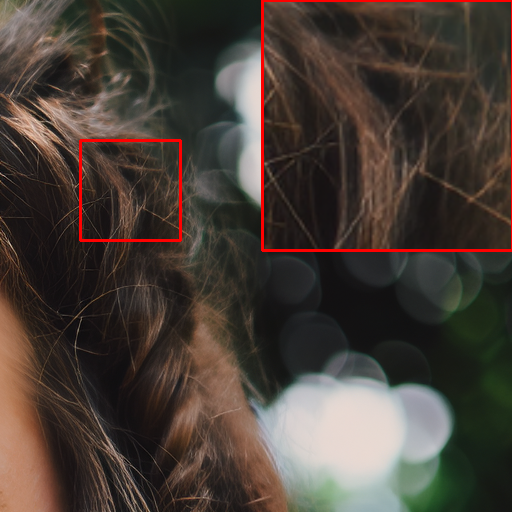}
        \vspace{-1.25cm}
        \begin{center}
        {\hspace{2.25cm}\colorbox{gray}{\color{white}50.42}}
        \end{center}
    \end{minipage}
    \begin{minipage}[t]{0.19\linewidth}
        \includegraphics[width=\linewidth]{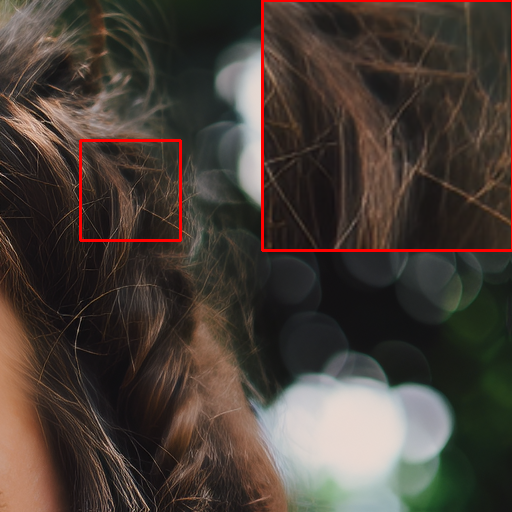}
        \vspace{-1.25cm}
        \begin{center}
        {\hspace{2.25cm}\colorbox{gray}{\color{white}53.28}}
        \end{center}
    \end{minipage}
    \begin{minipage}[t]{0.19\linewidth}
        \includegraphics[width=\linewidth]{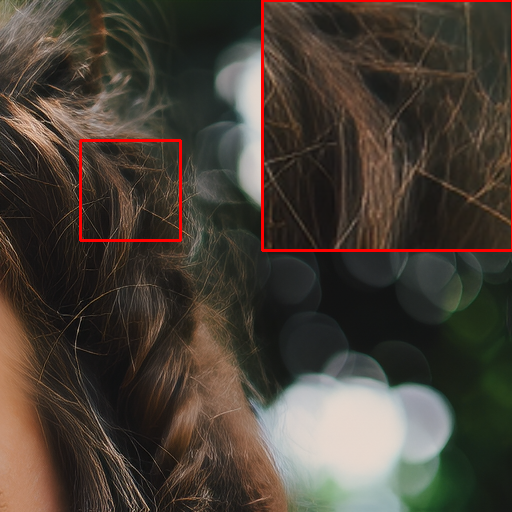}
        \vspace{-1.25cm}
        \begin{center}
        {\hspace{2.25cm}\colorbox{gray}{\color{white}54.70}}
        \end{center}
    \end{minipage}
    
    \vfill
    \begin{minipage}[t]{0.19\linewidth}
    \centering{LR}
    \end{minipage}%
    \begin{minipage}[t]{0.20\linewidth}
    \centering{Original GT}
    \end{minipage}%
    \begin{minipage}[t]{0.19\linewidth}
    \centering{SwinIR-UPos}
    \end{minipage}%
    \begin{minipage}[t]{0.20\linewidth}
    \centering{SwinIR-AMO}
    \end{minipage}%
    \begin{minipage}[t]{0.19\linewidth}
    \centering{SwinIR-AMO+FT}
    \end{minipage}
    \vfill
    \begin{minipage}[t]{0.19\linewidth}
        \includegraphics[width=\linewidth]{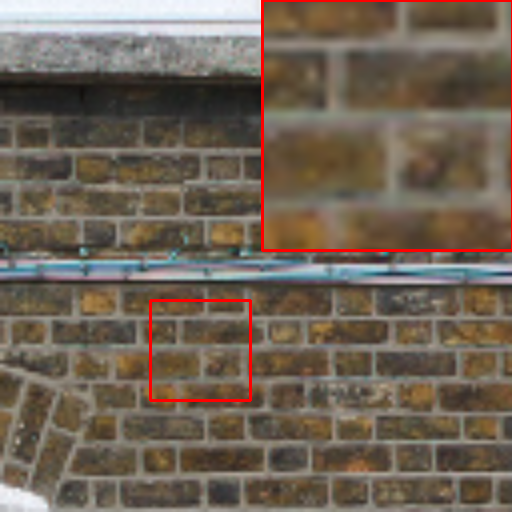}
        \vspace{-1.25cm}
        \begin{center}
            {\hspace{2.25cm}\colorbox{gray}{\color{white}35.03}}
        \end{center}
    \end{minipage}
    \begin{minipage}[t]{0.19\linewidth}
        \includegraphics[width=\linewidth]{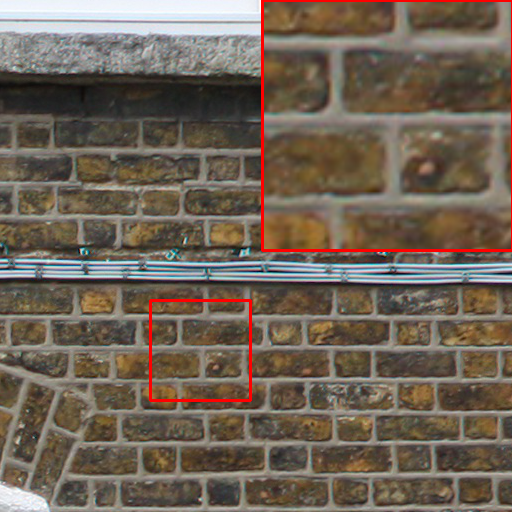}
        \vspace{-1.25cm}
        \begin{center}
        {\hspace{2.25cm}\colorbox{gray}{\color{white}48.67}}
        \end{center}
    \end{minipage}
    \begin{minipage}[t]{0.19\linewidth}
        \includegraphics[width=\linewidth]{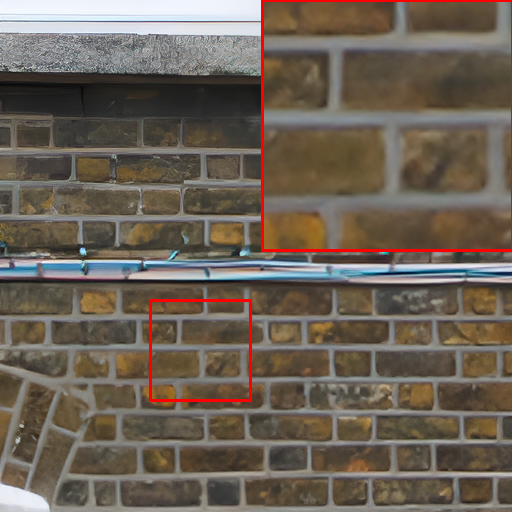}
        \vspace{-1.25cm}
        \begin{center}
        {\hspace{2.25cm}\colorbox{gray}{\color{white}50.48}}
        \end{center}
    \end{minipage}
    \begin{minipage}[t]{0.19\linewidth}
        \includegraphics[width=\linewidth]{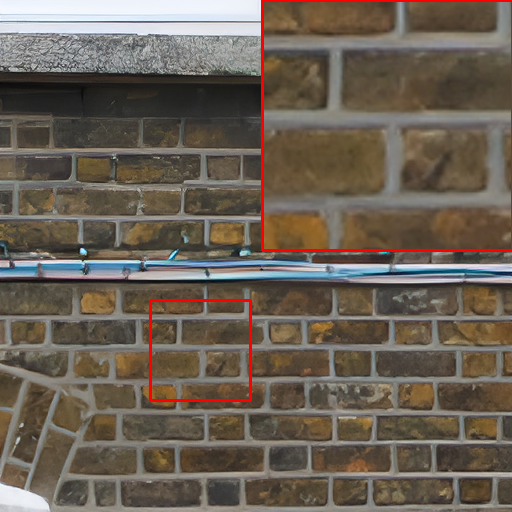}
        \vspace{-1.25cm}
        \begin{center}
        {\hspace{2.25cm}\colorbox{gray}{\color{white}57.30}}
        \end{center}
    \end{minipage}
    \begin{minipage}[t]{0.19\linewidth}
        \includegraphics[width=\linewidth]{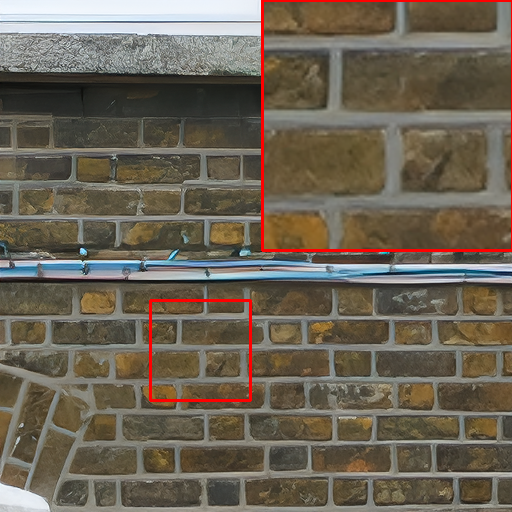}
        \vspace{-1.25cm}
        \begin{center}
        {\hspace{2.25cm}\colorbox{gray}{\color{white}65.22}}
        \end{center}
    \end{minipage}
    
    \vfill
    \begin{minipage}[t]{0.19\linewidth}
        \includegraphics[width=\linewidth]{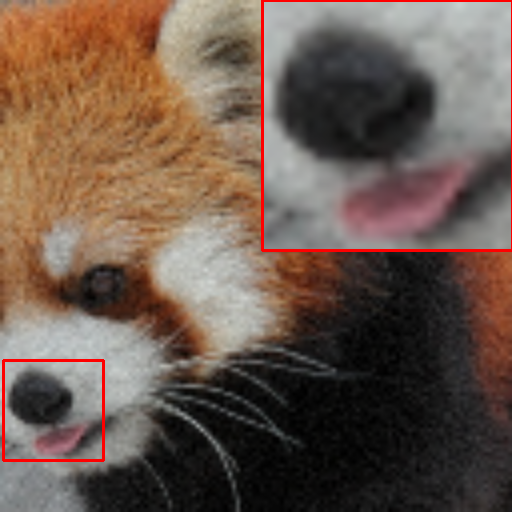}
        \vspace{-1.25cm}
        \begin{center}
        {\hspace{2.25cm}\colorbox{gray}{\color{white}47.17}}
        \end{center}
    \end{minipage}
    \begin{minipage}[t]{0.19\linewidth}
        \includegraphics[width=\linewidth]{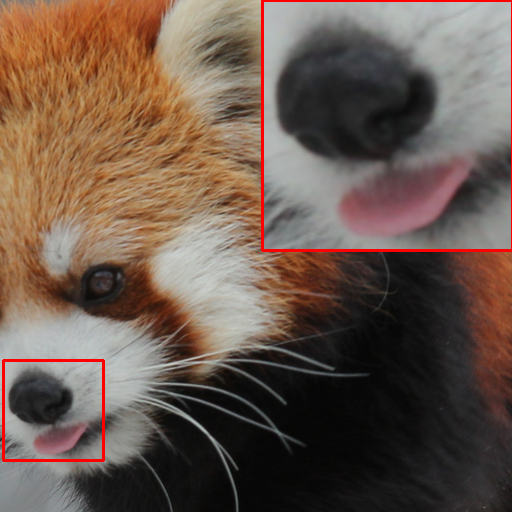}
        \vspace{-1.25cm}
        \begin{center}
        {\hspace{2.25cm}\colorbox{gray}{\color{white}71.16}}
        \end{center}
    \end{minipage}
    \begin{minipage}[t]{0.19\linewidth}
        \includegraphics[width=\linewidth]{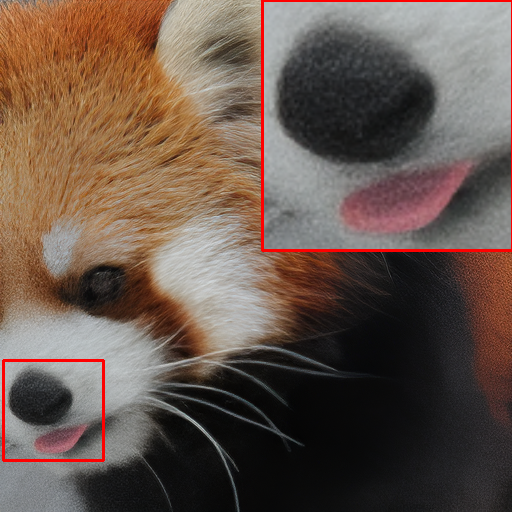}
        \vspace{-1.25cm}
        \begin{center}
        {\hspace{2.25cm}\colorbox{gray}{\color{white}71.10}}
        \end{center}
    \end{minipage}
    \begin{minipage}[t]{0.19\linewidth}
        \includegraphics[width=\linewidth]{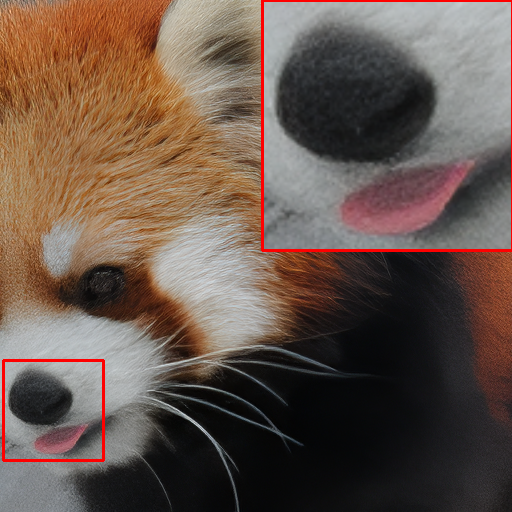}
        \vspace{-1.25cm}
        \begin{center}
        {\hspace{2.25cm}\colorbox{gray}{\color{white}74.23}}
        \end{center}
    \end{minipage}
    \begin{minipage}[t]{0.19\linewidth}
        \includegraphics[width=\linewidth]{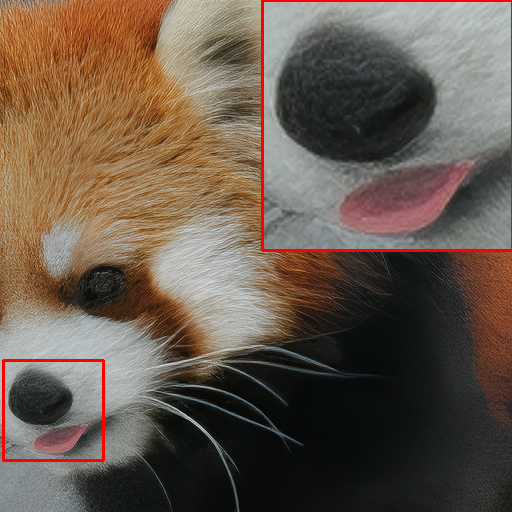}
        \vspace{-1.25cm}
        \begin{center}
        {\hspace{2.25cm}\colorbox{gray}{\color{white}76.27}}
        \end{center}
    \end{minipage}

    \vfill
    \begin{minipage}[t]{0.19\linewidth}
        \includegraphics[width=\linewidth]{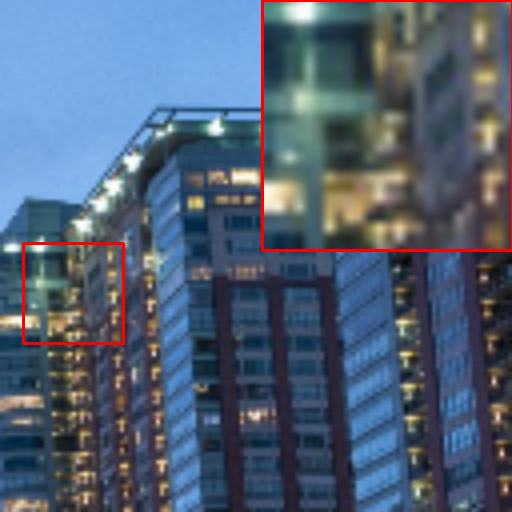}
        \vspace{-1.25cm}
        \begin{center}
        {\hspace{2.25cm}\colorbox{gray}{\color{white}41.69}}
        \end{center}
    \end{minipage}
    \begin{minipage}[t]{0.19\linewidth}
        \includegraphics[width=\linewidth]{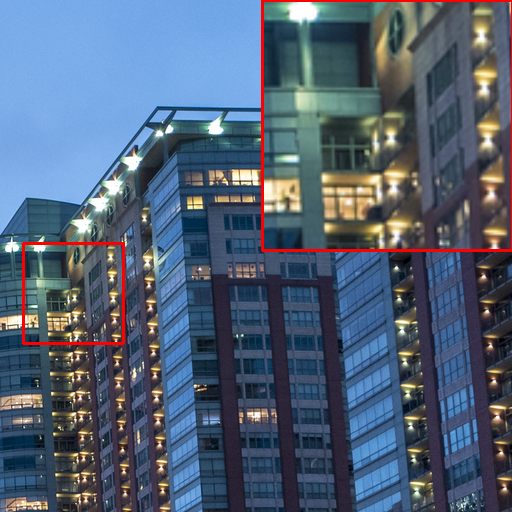}
        \vspace{-1.25cm}
        \begin{center}
        {\hspace{2.25cm}\colorbox{gray}{\color{white}68.72}}
        \end{center}
    \end{minipage}
    \begin{minipage}[t]{0.19\linewidth}
        \includegraphics[width=\linewidth]{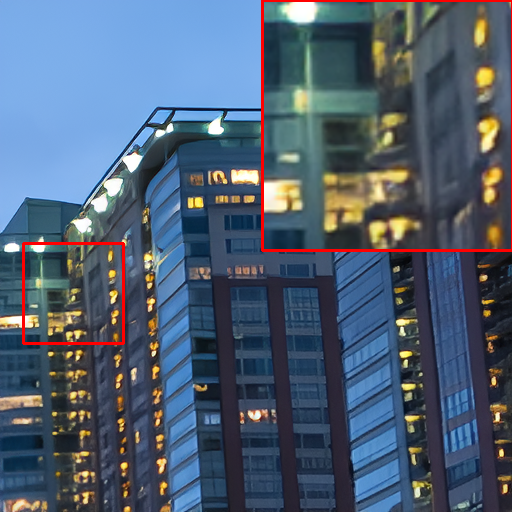}
        \vspace{-1.25cm}
        \begin{center}
        {\hspace{2.25cm}\colorbox{gray}{\color{white}72.06}}
        \end{center}
    \end{minipage}
    \begin{minipage}[t]{0.19\linewidth}
        \includegraphics[width=\linewidth]{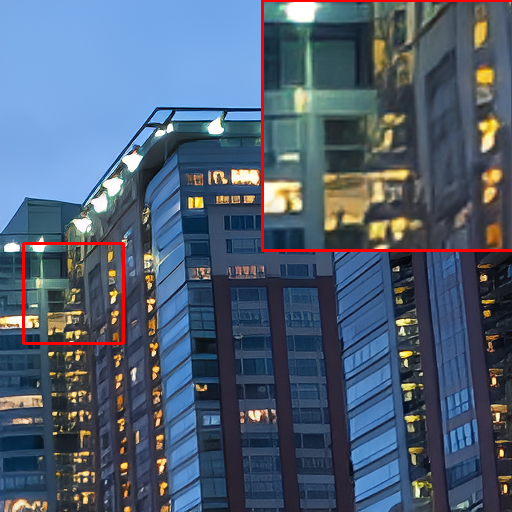}
        \vspace{-1.25cm}
        \begin{center}
        {\hspace{2.25cm}\colorbox{gray}{\color{white}73.53}}
        \end{center}
    \end{minipage}
    \begin{minipage}[t]{0.19\linewidth}
        \includegraphics[width=\linewidth]{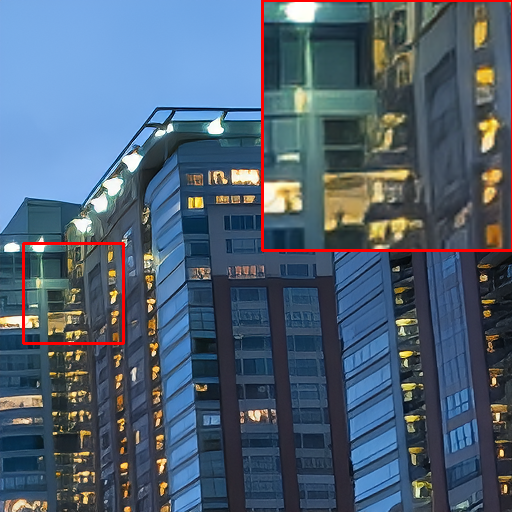}
        \vspace{-1.25cm}
        \begin{center}
        {\hspace{2.25cm}\colorbox{gray}{\color{white}74.73}}
        \end{center}
        
    \end{minipage}
    \begin{minipage}[t]{0.19\linewidth}
    \centering{LR}
    \end{minipage}%
    \begin{minipage}[t]{0.20\linewidth}
    \centering{Original GT}
    \end{minipage}%
    \begin{minipage}[t]{0.19\linewidth}
    \centering{RESRGAN-UPos}
    \end{minipage}%
    \begin{minipage}[t]{0.20\linewidth}
    \centering{RESRGAN-AMO}
    \end{minipage}%
    \begin{minipage}[t]{0.19\linewidth}
    \centering{RESRGAN-AMO+FT}
    \end{minipage}
    \\
    \caption{
        \textbf{Qualitative results with NR-IQA guidance.}
        Following the notation of Table~\ref{tab:mainresults},
        columns 3-5 are (top 3 rows) SwinIR-UPos, SwinIR-AMO, and SwinIR-AMO + FT, and (bottom 3 rows) Real-ESRGAN-UPos, Real-ESRGAN-AMO, and Real-ESRGAN-AMO + FT. 
        We show MUSIQ scores in insets.
        Qualitatively, we see improved performance as we move across the `UPos', `AMO', and `AMO-FT' methods, 
            particularly in terms of sharpness and detail generation.
        Zoom in for details.
    }
    \vspace{-0.3cm}
    \label{fig:suppqual}
\end{figure*}

Additional comparisons are shown in Fig.~\ref{fig:suppqual} (as in Fig.~\ref{fig:qual}).
Our method (AMO or AMO+FT) is universally sharper and more detailed than UPos 
(e.g., see the hair in row three).
Further, it can occasionally remove some of the noise present in the UPos scenario (see the tongue of the red panda). 
Importantly, our approach may not generate details that are identical to the GT, but it {does} construct sharp image content without jarring unrealistic artifacts 
(e.g., see rows one and four; the plants, rocks, and bricks have slightly different details, but they are plausible and of similar aesthetic quality nonetheless). %

\subsection{Additional Naive Optimization Visualizations}
\label{supp:sec:optart}
In Fig.~\ref{fig:supp:advartifacts},
as in Fig.~\ref{fig:advartifacts}, we show the subtle ``grid-like'' artifacts that appear when naive NR-IQA optimization is performed.
In particular, we see spatial patterns form in homogeneous areas (e.g., stripes in the sky or on the tan coloured island), while other areas exhibit highly unnatural colours 
(e.g., the alternating blue-red pixels on the dark rock).
These small, pixel-scale artifacts are akin to an adversarial attack on MUSIQ; hence, much of this structured noise is alleviated by applying LoRA (right insets).
Other methods of handling such artifacts, such as an adversarially robust NR-IQA model, may also be effective, but we leave this to future work.

\begin{figure}
    \centering
    \begin{minipage}[t]{0.9\linewidth}
    \includegraphics[width=\linewidth]{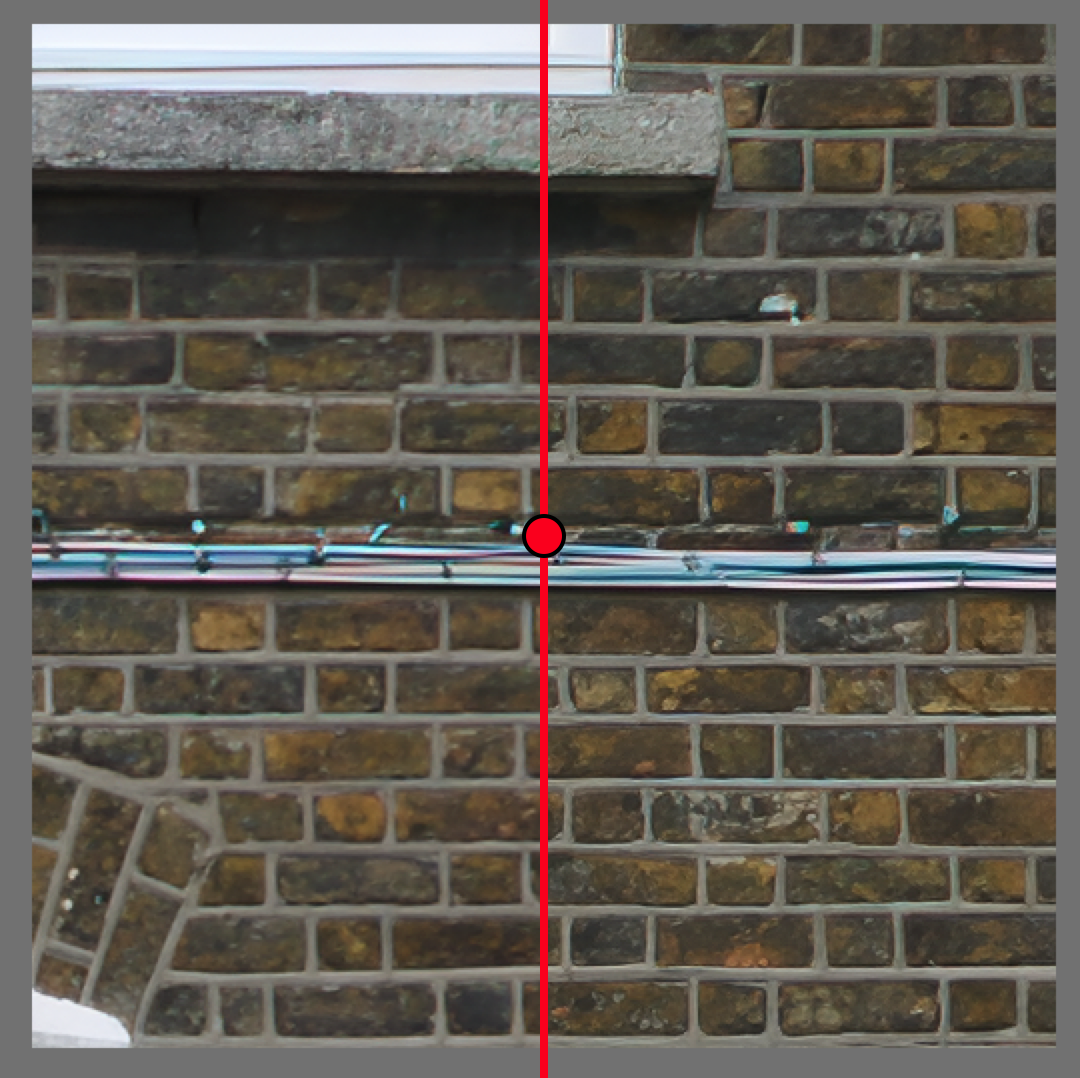}
    \end{minipage}
    \hfill
    \caption{
        \textbf{User study example.}
        Users can move the slider to alternate between 2 images.
    }
    \label{fig:demo}
\end{figure}

\section{Additional Results on RealSR}
\label{supp:sec:realsr}
We provide results on the 
RealSRv3 \cite{cai2019toward} dataset in
Table \ref{tab:resultsr}.
Similar to the HGGT test dataset, we find that \textit{our method is superior in terms of every NR-IQA metric}, at the expense of the exact pixel-level details measured by PSNR and SSIM (following the perception-distortion tradeoff \cite{blau2018perception}). However, according to mid-level {FR} metrics, our method \textit{also} performs well, obtaining the best scores on LPIPS-ST, and even on LPIPS and DISTS for SwinIR. This suggests our method can improve image quality, while maintaining the most salient perceptual details (e.g., mid-level textures) of the underlying GT.

\section{Comparative Evaluation via User Study}
\label{supp:sec:userstudy}
Similar to the HGGT user study, we invite 12 volunteers to evaluate their preference between SwinIR-AMO+FT and SwinIR-UPos, using the HGGT Test-100 dataset. 
Each volunteer evaluates 25 image pairs (25\% of the dataset), with each image in Test-100 being seen an equal number of times (namely, three). 
For each pair (SwinIR-AMO+FT vs.\ SwinIR-UPos), we employ an image comparison slider. 
This tool places two images on top of each other, and allows volunteers to use a slider to alternate between them (see Fig.~\ref{fig:demo} for a visualization). 
The order of presentation of the two methods (left vs.\ right) is randomized to eliminate bias.
For each individual, we obtain a single score, which is the percentage of the time that they prefer our method (across those 25 images).
The average score across raters is \textbf{69.7\%} (median: 68.0\%; empirical standard error of the mean: 4.8\%), suggesting our algorithm is preferred over the HGGT-based UPos approach at a more than 2:1 ratio, \textit{despite their use of human annotations}, which ours does not use.
Following similar image quality assessment protocols (e.g., \cite{series2012methodology}), a simple single-sample one-sided t-test finds the rater mean significantly above 50\% ($p < 0.01$; 95\% confidence interval: $[60.2\%,79.1\%]$). %

\section{Remark on Evaluation Metric Types and Nomenclature}
\label{supp:metric remark}
The perception-distortion tradeoff \cite{blau2018perception} necessitates a complex suite of evaluation metrics that consider different aspects of the SR outputs, including pixel-level fidelity to a GT image and standalone image quality.
Some works (e.g., \cite{wu2024seesr}) even utilize performance on downstream vision tasks (e.g., detection or segmentation) as a form of checking semantic preservation.
In this work, we therefore also include a continuum of metrics, which we hope will cover various points along the perception-distortion frontier.
These metrics are often categorized along two different axes: (i) the use of a reference and (ii) the level of visual abstraction (low vs mid vs high).

\noindent
\textbf{NR vs FR.} The first form of metric categorization is full-reference (FR) vs no-reference (NR). In general, FR metrics (which have access to a GT) measure distortion, while NR metrics (which do not use a GT) measure perceptual quality. 
For NR metrics, there is no way to measure distortion; however, there are many different aspects of perceptual quality that can be considered, ranging from simple sharpness to differentiating aesthetic vs technical quality (e.g., \cite{wu2023exploring}). Hence, it is common (e.g., \cite{wu2024seesr,wu2024one,yang2024pixel}) to use a set of NR-IQA models, which presumably complement each other, as we do (see \S\ref{sec:analysis} and \S\ref{supp:sec:nriqachoices} for the discussion behind our metric choices).
For FR metrics, there is more of a spectrum (i.e., they can include some aspects of perceptual information, in addition to measuring distortion).
PSNR and other per-pixel distances have no notion of perception, operating directly on pixel values.
SSIM is meant to be more perceptual, but is a simple, hand-crafted similarity operating on colours, limiting its perceptual modelling capabilities \cite{nilsson2020understanding}.
In contrast, LPIPS and DISTS utilize neural network features, aiming to capture certain aspects of human vision. 
They are therefore \textit{more perceptual} than, e.g., PSNR, as they will tolerate some pixel differences (distortion) if they improve network activation similarity.
Even further along this curve towards greater perceptual sensitivity is LPIPS-ST, a model designed specifically to ignore small spatial shifts (which are devastating to pixel-level distortion measures).
Indeed, in many cases, we find that LPIPS-ST actually agrees with the NR-IQA perceptual metrics more closely than LPIPS or DISTS,
despite being an FR metric.
Hence, FR metrics can occupy a range across the perception-distortion curve.

\noindent
\textbf{Abstraction Level.}
A separate nomenclature arises based on the \textit{type of information} that impacts the model.
It is based on the hierarchical nature of \textit{biological} vision
(e.g., \cite{peirce2015understanding}), 
but is also commonly used throughout computer vision
(e.g., \cite{fu2023dreamsim}). %
Specifically, we divide visual processes into 
\textit{low-level}, 
    relating to raw colours and 2D geometry (e.g., edges);
\textit{mid-level},
    encompassing ``groupings'' of more basic features into patterns and textures, as well as local 3D structures;
and \textit{high-level},
    pertaining to semantics (e.g., scene classification) and representational abstraction (e.g., holistic interpretations of the image).
For this reason, we refer to PSNR and SSIM, which operate directly on colours, as low-level, while LPIPS and DISTS are mid-level, as they respond best to textures, image ``styles'', and other regional ``grouped'' visual elements.
We label neural NR-IQA models, such as MUSIQ, as high-level, as they process the image holistically, taking semantic context into account, as well as aesthetics, though they may also care about low-level issues, such as noise and blur. 
In general, including in our work, low-level metrics tend to measure distortion, while mid-level and high-level ones are more related to perceptual quality. 
However, there may be exceptions: 
for instance, measuring sharpness via a simple image filter is a low-level NR metric that targets perceptual quality rather than distortion 
(e.g., \cite{tolstikhin2017wasserstein}).

\section{Limitations}
\label{supp:limits}

\begin{figure}
    \centering
    \begin{minipage}[t]{0.3\linewidth}
        \includegraphics[width=\linewidth]{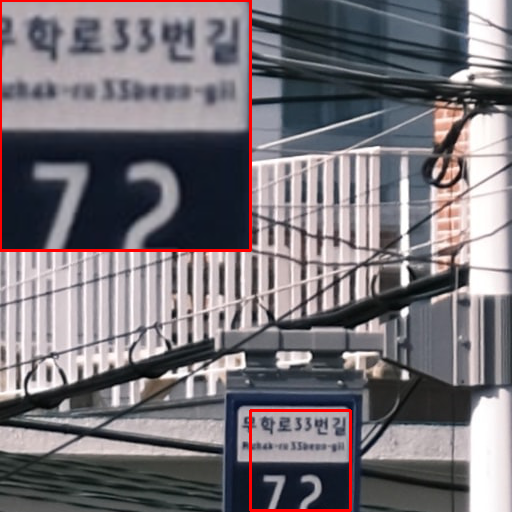}
        \vspace{-1.25cm}
        \begin{center}
        {\hspace{2.25cm}\colorbox{gray}{\color{white}48.88}}
        \end{center}
    \end{minipage}
    \begin{minipage}[t]{0.3\linewidth}
        \includegraphics[width=\linewidth]{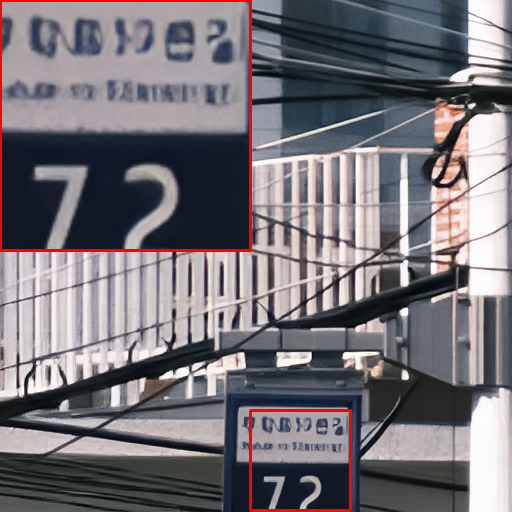}
        \vspace{-1.25cm}
        \begin{center}
        {\hspace{2.25cm}\colorbox{gray}{\color{white}61.00}}
        \end{center}
    \end{minipage}
    \begin{minipage}[t]{0.3\linewidth}
        \includegraphics[width=\linewidth]{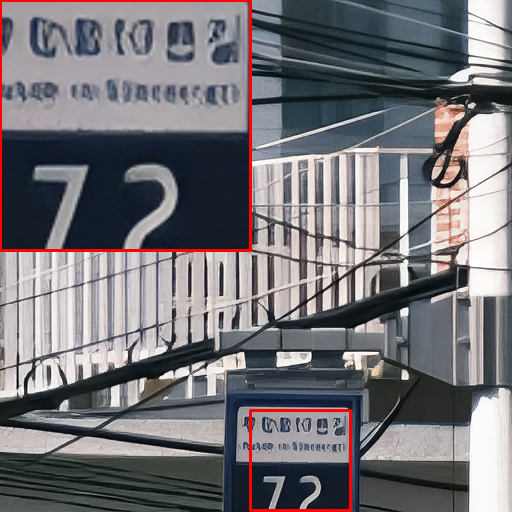}
        \vspace{-1.25cm}
        \begin{center}
        {\hspace{2.25cm}\colorbox{gray}{\color{white}75.24}}
        \end{center}
    \end{minipage}
    
    \vfill
    \begin{minipage}[t]{0.3\linewidth}
    \centering{Original GT}
    \end{minipage}
    \begin{minipage}[t]{0.3\linewidth}
    \centering{RESRGAN-UPos}
    \end{minipage}%
    \begin{minipage}[t]{0.3\linewidth}
    \centering{RESRGAN-AMO+FT}
    \end{minipage}%
    \vfill
    
    \begin{minipage}[t]{0.3\linewidth}
        \includegraphics[width=\linewidth]{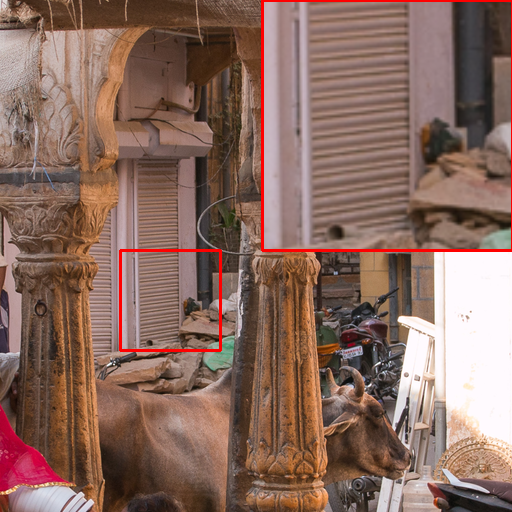}
        \vspace{-1.25cm}
        \begin{center}
            {\hspace{2.25cm}\colorbox{gray}{\color{white}73.50}}
        \end{center}
    \end{minipage}
    \begin{minipage}[t]{0.3\linewidth}
        \includegraphics[width=\linewidth]{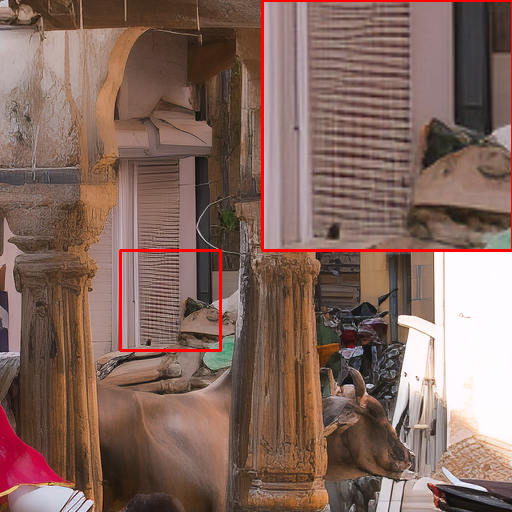}
        \vspace{-1.25cm}
        \begin{center}
        {\hspace{2.25cm}\colorbox{gray}{\color{white}76.39}}
        \end{center}
    \end{minipage}
    \begin{minipage}[t]{0.3\linewidth}
        \includegraphics[width=\linewidth]{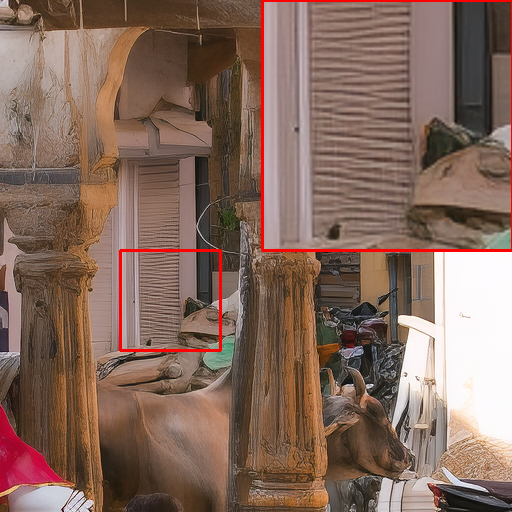}
        \vspace{-1.25cm}
        \begin{center}
        {\hspace{2.25cm}\colorbox{gray}{\color{white}77.70}}
        \end{center}
    \end{minipage}
    
    \begin{minipage}[t]{0.3\linewidth}
    \centering{Original GT}
    \end{minipage}%
    \begin{minipage}[t]{0.3\linewidth}
    \centering{SwinIR-UPos}
    \end{minipage}%
    \begin{minipage}[t]{0.3\linewidth}
    \centering{SwinIR-AMO+FT}
    \end{minipage}%
    \\
    \caption{
        \textbf{Illustration of limitations.}
        We show examples of shortcomings of our method (see Fig.~\ref{supp:limits}),
            with MUSIQ scores in insets.
        In row one, we show the shortcomings of our model with respect to text, a particularly difficult form of image content.
        In row two, we see that our model does still incur artifacts, such as the mangled lines in the zoomed inset.
    }
    \vspace{-0.3cm}
    \label{fig:limitation}
\end{figure}

While our IQA-based method is able to sharpen SR outputs, as well as hallucinate aesthetically pleasing details in most cases, there are still several shortcomings to our approach.
First, higher IQA model score does not guarantee improved human perceptual quality nor does it strictly ensure our outputs are artifact-free.
This is related to the discussion in \S\ref{sec:methods:do} and Fig.~\ref{fig:advartifacts}, 
    where we postulate that some image changes can improve IQA score despite worsening perceptual quality
    (e.g., direct optimization being similar to an adversarial attack on the quality model).
In Fig.~\ref{fig:limitation}, row two, for instance, we see that the SR model fails to predict the correct image details, 
    leading to incorrect line orientations and aliasing-like artifacts 
    (though the UPos baseline in column two arguably has worse artifacts).
Second, from a semantic perspective, certain classes of image content may require different treatment, 
     the requirements of which NR-IQA models are not naturally aware.
For example, row one in Fig.~\ref{fig:limitation} demonstrates how super-resolved \textit{text} can become mangled.
In terms of human preference, it can be argued that having a blurrier output in such uncertain cases may be more desirable (i.e., having blurred characters, rather than \textit{wrong} characters, could be preferred for text).
Nevertheless, text is notoriously challenging to super-resolve (prompting development of specialized methods for it \cite{li2023learning,zhang2024diffusion}); further, the UPos baseline suffers from similar artifacts as our outputs.
Overall, we suspect better IQA models or more sophisticated regularized optimizations (i.e., beyond LoRA) can mitigate some of the artifacts incurred by our approach. Handling more semantic issues, such as text hallucination, may require more specialized models.

\end{document}